\documentclass[journal]{IEEEtran}
\usepackage{graphicx}
\usepackage{mathptmx}
\usepackage{latexsym}
\usepackage{color}
\usepackage{gensymb}
\usepackage{multirow}
\usepackage{times}
\usepackage{epsfig}
\usepackage{amsmath}
\usepackage{amssymb}
\usepackage{array}
\usepackage{algpseudocode}
\usepackage{mwe}
\usepackage{cite}
\usepackage{multicol}
\usepackage{romannum}
\usepackage{tocbibind}
\usepackage[font = scriptsize]{caption}
\usepackage[labelformat = simple]{subfig}
\usepackage[norelsize,linesnumbered,longend,ruled,lined,boxed,commentsnumbered]{algorithm2e}

\begin{document}

\title{Local Jet Pattern: A Robust Descriptor for Texture Classification}

\author{  Swalpa~Kumar~Roy\IEEEauthorrefmark{2},
         Bhabatosh~Chanda,
         Bidyut~B.~Chaudhuri,
%		 Soumitro~Banerjee,    
         Dipak~Kumar~Ghosh
         and~Shiv~Ram~Dubey
         \thanks{ 

\indent S. K. Roy and B. Chanda are with the Electronics and Communication Sciences Unit at Indian Statistical Institute, Kolkata 700108, India (email: swalpa@ieee.org; chanda@isical.ac.in).
\newline 
\indent B. B. Chaudhuri is with the Computer Vision and Pattern Recognition Unit at Indian Statistical Institute, Kolkata 700108, India (email: bbc@isical.ac.in).
\newline
%\indent S. Banerjee is with the Department of Physical Sciences at Indian Institute of Science Education and Research, Kolkata, Mohanpur Campus, 741246, India (email: soumitro@iiserkol.ac.in). \newline
\indent D. K. Ghosh is with the Department of Electronics \& Communication Engineering at National Institute of Technology, Rourkela, Orissa 769008, India (email: dipak@ieee.org).
\newline
\indent S. R. Dubey is with the Computer Vision Group at Indian Institute of Information Technology, Sri City, Andhra Pradesh-517646, India (email: srdubey@iiits.in).
\newline
\IEEEauthorrefmark{1} Corresponding author
% phone: (91) (33) 2575-2852, fax: (91) (33) 2577 3035 
email: swalpa@students.iiests.ac.in, swalpa@ieee.org.
}}

\markboth{Preprint: Accepted in Multimedia Tools and Applications, Springer}%
{Dubey \MakeLowercase{\textit{et al.}}: Bare Demo of IEEEtran.cls for IEEE Journals}
\maketitle

\begin{abstract}
Methods based on locally encoded image features have recently become popular for texture classification tasks, particularly in the existence of large intra-class variation due to changes in illumination, scale, and viewpoint. Inspired by the theories of image structure analysis, this work proposes an efficient, simple, yet robust descriptor namely local jet pattern~(\textsc{Ljp}) for texture classification. In this approach, a jet space representation of a texture image is computed from a set of derivatives of Gaussian~(DtGs) filter responses up to second order, so-called local jet vectors~(\textsc{Ljv}), which also satisfy the Scale Space properties. The \textsc{Ljp} is obtained by using the relation of center pixel with its' local neighborhoods in jet space. Finally, the feature vector of a texture image is formed by concatenating the histogram of \textsc{Ljp} for all elements of \textsc{Ljv}. All DtGs responses up to second order together preserves the intrinsic local image structure, and achieves invariance to scale, rotation, and reflection. This allows us to design a discriminative and robust framework for texture classification. Extensive experiments on five standard texture image databases, employing nearest subspace classifier~(\textsc{Nsc}), the proposed descriptor achieves 100\%, 99.92\%, 99.75\%, 99.16\%, and 99.65\% accuracy for Outex\_TC10, Outex\_TC12, KTH-TIPS, Brodatz, CUReT, respectively, which are better compared to state-of-the-art methods.
\end{abstract}

\section{Introduction}
\label{sec:intro}
% The very first letter is a 2 line initial drop letter followed by the rest of the first word in caps.
Computer vision and pattern classification areas have seen enormous progress in the last five decades, having several potential applications such as multimedia event detection~\cite{ChangMa2017}, texture recognition~\cite{ojala2002multiresolution,guo2010completed}, motion detection~\cite{Chang2017}, event analysis~\cite{ChangX2017}, face and head pose recognition~\cite{Li2017}, image retrieval \cite{zhu2014weighting}, mobile landmark search \cite{zhu2016learning}, mobile and multimedia retrieval \cite{ zhu2015topic, xie2016unsupervised} etc.
% \IEEEPARstart{T}{exture} is 
Since texture is a repeated basic primitives visual pattern that describes the distinctive appearance of many natural objects and is present almost everywhere,
% . Texture classification 
it shows a useful role in the areas of computer vision and pattern classifications namely object recognition, natural scene identification, material classification,  and image segmentation~\cite{xie2008galaxy,dubey2015multi,umer2017novel}. A crucial issue of texture classification is efficient texture representation, which can be categorized into geometrical, statistical,  structural, signal processing based, and model-based methods~\cite{haralick1973textural, randen1999filtering,dubey2017local}. 
A picture may be captured under various geometric and photo-metric varying conditions, and hence an ideal texture model for texture classification should be robust enough against the changes in illuminance, view-point, rotation or reflection, scale, and geometry of the underlying surface. Earlier texture classification methods mainly concentrated on the statistical analysis of texture images. In recent literature, two categories of texture classification approaches dominate current research, namely texture descriptor based methods~\cite{ojala2002outex, guo2010completed, quan2014distinct} and deep learning based methods \cite{krizhevsky2012imagenet, cimpoi2015deep, cimpoi2016deep, andrearczyk2016using, liu2016evaluation}. In deep learning based methods a Convolutional Neural Network (\textsc{Cnn}) has been proposed and trained to classify the texture images~\cite{chan2015pcanet, cimpoi2016deep}. The deep learning based methods offer good classification performance, however, it has the following limitations: it requires a large amount of data and it is computationally expensive to train. The complex models take weeks to train sometimes using several machines equipped with expensive GPUs. Also, at present, no strong theoretical foundation of finding the topology/training method/flavor/hyper-parameters for deep learning exist in the literature. On the other hand, the descriptor based methods have the advantage of easy to use, data independence, and robustness to real-life challenges such as illumination and scale differences. Therefore, attention is given here to form the local texture descriptors which are adequate for achieving local  invariance~\cite{zhang2007local, crosier2010using,
varma2005statistical, varma2009statistical, zand2015texture}. 

The search for invariances started in the nineties of last century \cite{weiss1993geometric}. In the beginning, a circular autoregressive dense model has been proposed by Kashyap and Khotanzad to study the rotation invariant texture classification~\cite{kashyap1986model}. To study rotation invariance texture classification, many other models were explored such as  hidden Markov model~\cite{chen1994rotation}, multi-resolution~\cite{mao1992texture}, Gaussian Markov model~\cite{deng2004gaussian}, etc. Currently, Varma and Zisserman introduced a texton dictionary-based method for rotation invariant texture classification~\cite{varma2005statistical}. Later on, they also proposed another texton based method where the features are directly represented utilizing the image local patch~\cite{varma2009statistical}. Recently, some works are also introduced scale and affine invariant feature extraction for texture classification. Chaudhuri and Sarkar proposed a technique for texture recognition based on differential box-counting~(\textsc{Dbc}) \cite{chaudhuri1995texture} based algorithm. Varma and Garg~\cite{varma2007locally} extracted a local fractal vector for each pixel, and computed a statistical histogram; Liu and Fieguth~\cite{liu2012texture} applied random projection for densely sampled image patches, and extracted the histogram signature; Yao and Sun~\cite{yao2003retrieval} normalized statistical edge feature distribution to resist variation in scale; Lazebinik \textit{et al.}~\cite{lazebnik2005sparse} and Zhang \textit{et al.}~\cite{zhang2007local} detected Harris and Laplacian regions and extracted texture signatures after normalizing these regions. Recently, global scale invariant feature extraction methods drew attention because local scale normalization is usually slow due to pixel by pixel operations. Xu \textit{et al.}~\cite{xu2009viewpoint} and Quan \textit{et al.}~\cite{quan2014distinct} have classified the image pixels into multiple point sets by gray intensities or local feature descriptors. Other than extracting scale invariant features, pyramid histograms with shifting matching~\cite{crosier2010using} scheme were also proposed by some researchers~\cite{zhang2013local}. In order to impart more robustness, feature extraction is often performed locally. In 1996 a computationally efficient texture descriptor, called local binary pattern~(\textsc{Lbp}) was proposed by Ojala \textit{et al.} \cite{ojala1996comparative, ojala2002multiresolution} for gray-scale and rotation invariant texture classification. Later on, other variants of \textsc{Lbp} such as center-symmetric \textsc{Lbp}~(\textsc{CsLbp})~\cite{heikkila2009description}, derivative-based \textsc{Lbp}~\cite{zhang2010local}, \textsc{Lbp} variance (\textsc{Lbpv})~\cite{guo2010rotation}, the dominant \textsc{Lbp} (\textsc{Dlbp})~\cite{liao2009dominant}, the completed model of \textsc{Lbp}~(\textsc{Clbp})
\cite{guo2010completed}, order-based local descriptor~\cite{dubey2014rotation}, local wavelet pattern~\cite{dubey2015local}, multichannel decoded \textsc{Lbp}~\cite{dubey2016multichannel}, Complete Dual-Cross Pattern~(\textsc{Cdcp})~\cite{roy2017cdcp}, Local ZigZag Pattern~(\textsc{Lzp})~\cite{roy2018local}, and Local Morphological Pattern~(\textsc{Lmp})~\cite{roy2018lmp} etc. were introduced in numerous applications of computer vision and pattern recognition. But these methods could not address the scaling issue very well and \textsc{Lbp} feature alone could not achieve good performance. However, the feature values of a texture image vary with the scale, it is difficult to know if a query texture image has the same scale as the target images. Recently, Quan \textit{et al.}~\cite{quan2014distinct} proposed a multi-scale \textsc{Lbp} based global fractal feature to solve the scale issue. But this feature is not robust for small sized images. Li \textit{et al.} \cite{li2012scale} proposed a technique to find the optimal scale for each pixel and extracted \textsc{Lbp} feature with the optimal scale. However, it failed to extract accurate and consistent scale for all pixels. To extract scale invariant feature, Guo \textit{et al.}~\cite{guo2016robust} proposed scale selective complete local binary pattern-(\textsc{SsClbp}). Although their reported results are quite impressive, they did not establish the relation between image structure and scale parameter. 

Noise, which introduces additional geometric variation, can also create problems for texture image recognition. Hence, noise tolerant and robust texture recognition problem drew attention of the research communities. Tan and Triggs \cite{tan2007enhanced} proposed local ternary pattern~(\textsc{Ltp}) for noise invarint texture classification. However, \textsc{Ltp} is not purely gray-scale invariant. Ren \textit{et al.} \cite{ren2013noise} introduced a relativly more efective noise resistant local binary pattern~(\textsc{Nrlbp}) scheme based on \textsc{Lbp}, but it is computationaly expensive for bigger scales with larger number of neighboring points. Liu \textit{et al.} proposed binary rotation invariant and noise tolerant (\textsc{Brint}) feature \cite{liu2014brint}. However, to calculate \textsc{Brint}, they used a multi-scale approach which is computationally expensive and suffers from high dimensionality problem. To the best of our knowledge, no work is reported in the literature that simultaneously  deals with invariance to scale, translation, rotation, or reflection and insensitive to noise of a texture image. To achieve such invariance properties, it appears that local patterns from a Jet Space~\cite{florack1996gaussian} of the image may be useful. Based on this assumption, we propose here an effective local jet space descriptor, named as local jet pattern~(\textsc{Ljp}). To extract the \textsc{Ljp}, at first the Jet Space of a texture image is derived by convolving the derivative of Gaussian~(DtGs) kernels upto $2^{nd}$ order. Then, the \textsc{Ljp} for each element of Jet Space is built by exploiting the local relationships in that space. Finally, for a texture image, the normalized feature vector is formed by concatenating histogram of \textsc{Ljp} for all elements of the Jet Space. The main contributions of this paper can be summarized as follows,
\begin{itemize}

\item We propose a simple, effective, yet robust jet space based texture descriptor called local jet pattern~(Sec.~\ref{sec:prop}) for texture classification. In the jet space, the properties of Hermite polynomial are utilized to make DtGs responses for preserving the intrinsic local image structure in a hierarchical way.

\item The \textsc{Ljp} descriptor achieves sufficient invariance to address the challenges of scale, translation, and rotation (or reflection) for texture classification. In addition, the proposed descriptor is also robust to the variations of noise.

%\item We also showed that the jet space satisfies the properties of Scale Space (Subsec.~\ref{subsec:JSN}).
\item The proposed algorithm takes very reasonable amount of time in feature extraction.

\item We experimented and observed that the proposed descriptor outperforms to the traditional \textsc{Lbp} and other state-of-the-art methods on different benchmark texture databases, such as KTH-TIPS and CUReT (Sec.~\ref{sec:results}) etc.
\end{itemize}

The rest of the paper is organized as follows. Sec.~\ref{sec:LFJS} deals with the mathematical definition of a broad class of local image decomposition's, jet and jet space. The proposed \textsc{Ljp} feature extraction scheme is presented in Sec.~\ref{sec:prop}. In Sec.~\ref{sec:results} the performance of texture classification is compared with the state-of-the-art methods. Finally, concluding remarks are drawn in Sec.~\ref{sec:conclusion}.

\section{Image Structure Analysis, Jets and Jet Space}
\label{sec:LFJS}
Image analysis is an important part of most image processing tasks. %The aim of image analysis is to derived feature from visual data for further processing and analysis. In order to support recognition of several image structures embedded in image data and to serve as a precursor for more detailed analysis, such as defining a local orientation or scale, or labelling a region as ``corner'', ``edge'' etc., a local image structure analysis is demanded. 
As a general framework to deal with image structures at different resolution, the Scale Space representation is introduced in~\cite{lindeberg1994scale}. The idea of ``local deep structure" is linked to the concept of derivative: given by $$I'(0) = \lim_{\Delta \to 0}\frac{I(0 + \Delta) - I(0)}{\Delta} $$  where $\mathcal{I}$ denotes image intensity. This expression is not appropriate to discrete domain, such as images, that are the result of quantized measurement. The scale space analysis~\cite{lindeberg1994scale} introduces a two-step solution for image derivative measurement as follows: \\
\indent First, a measure of inner scale changing of an image is described, by convolving (denoted by $\ast$) the image with Gaussian kernels. The 1$\mathcal{D}$ Gaussian kernels at scale $\sigma \in \mathbb{R}^{+}$ are defined as
$$G_{\sigma}(x) = \frac{1}{\sigma\sqrt{2 \pi}} \exp^{\frac{-x^{2}}{2\sigma^{2}}} $$

\noindent The convolution of 2-$\mathcal{D}$ Gaussian function with an image patch can be smoothly computed by applying the 1-$\mathcal{D}$ Gaussian function using two passes in the horizontal and vertical directions which exhibits the separable~\cite{lowe2004distinctive} property of 2-$\mathcal{D}$ Gaussian function:
$ G_{\sigma}(x,y) = G_{\sigma}(x) G_{\sigma}(y). $
\noindent The scaling operation to obtain $\mathrm{ I_{\sigma} = G_{\sigma} \ast I}$ can be computed efficiently and stably due to the space and frequency localization of the Gaussian~\cite{victor2003simultaneously} even if the input image function ${I}$ is directly sampled and yeilds the result of physical measurement.
\begin{figure}[htp]
%\captionsetup{justification=left}
\begin{minipage}[b]{1.0\linewidth}
\centering
\centerline{\includegraphics[clip=true, trim=0  255 380 10, width=0.99\linewidth]{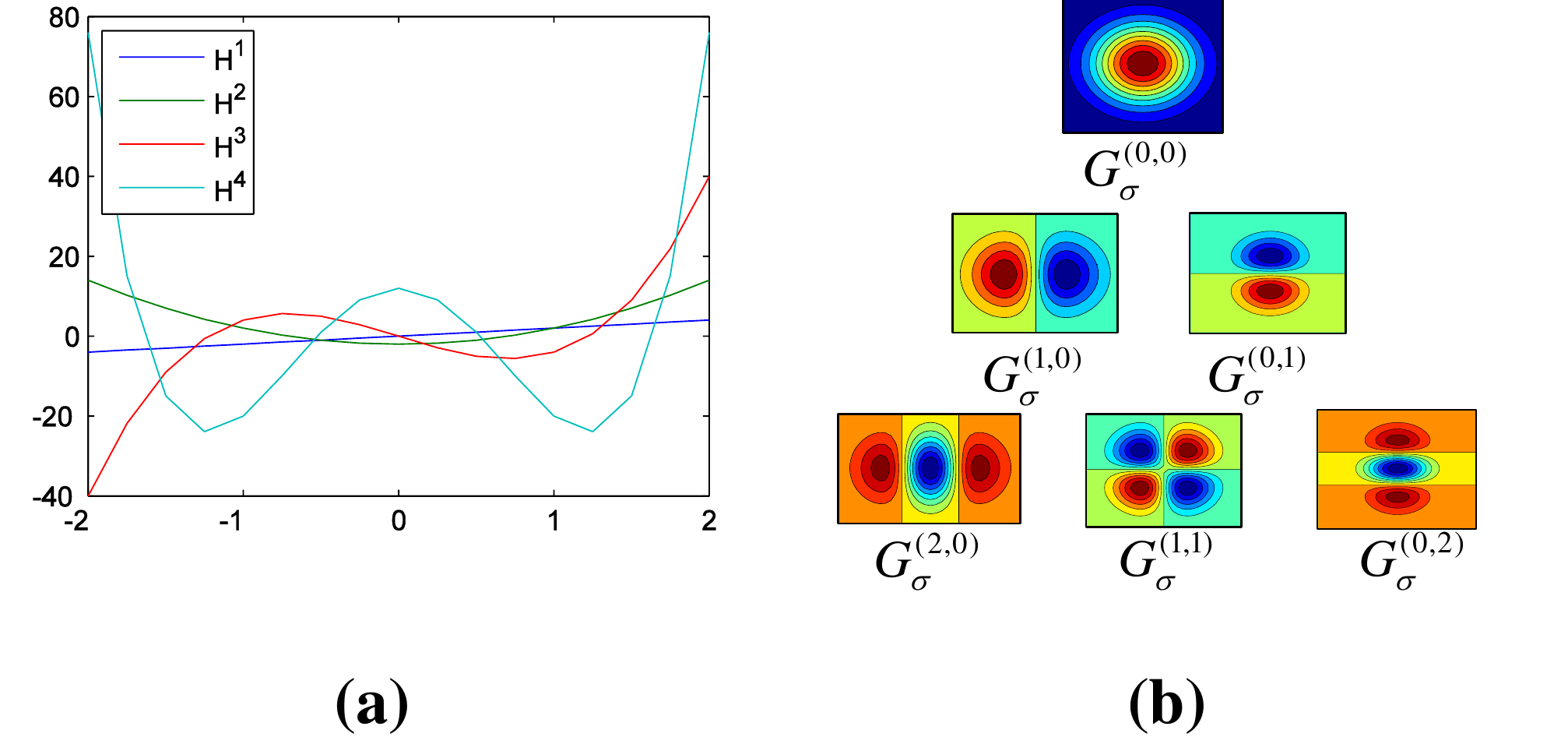}} %trim - left, bottom, right, top
\end{minipage}
\caption{(a) Hermite polynomial of different orders. (b) The 2-$\mathcal{D}$ derivatives of the Gaussian (DtGs) upto $2^{nd}$ order.}
\label{fig:Hermite_DtGs}
\end{figure}

\begin{figure}[htp]
%\captionsetup{justification=left}
\begin{minipage}[b]{1.0\linewidth}
\centering
\centerline{\includegraphics[clip=true, trim=80 170 260 20, width=0.95\linewidth]{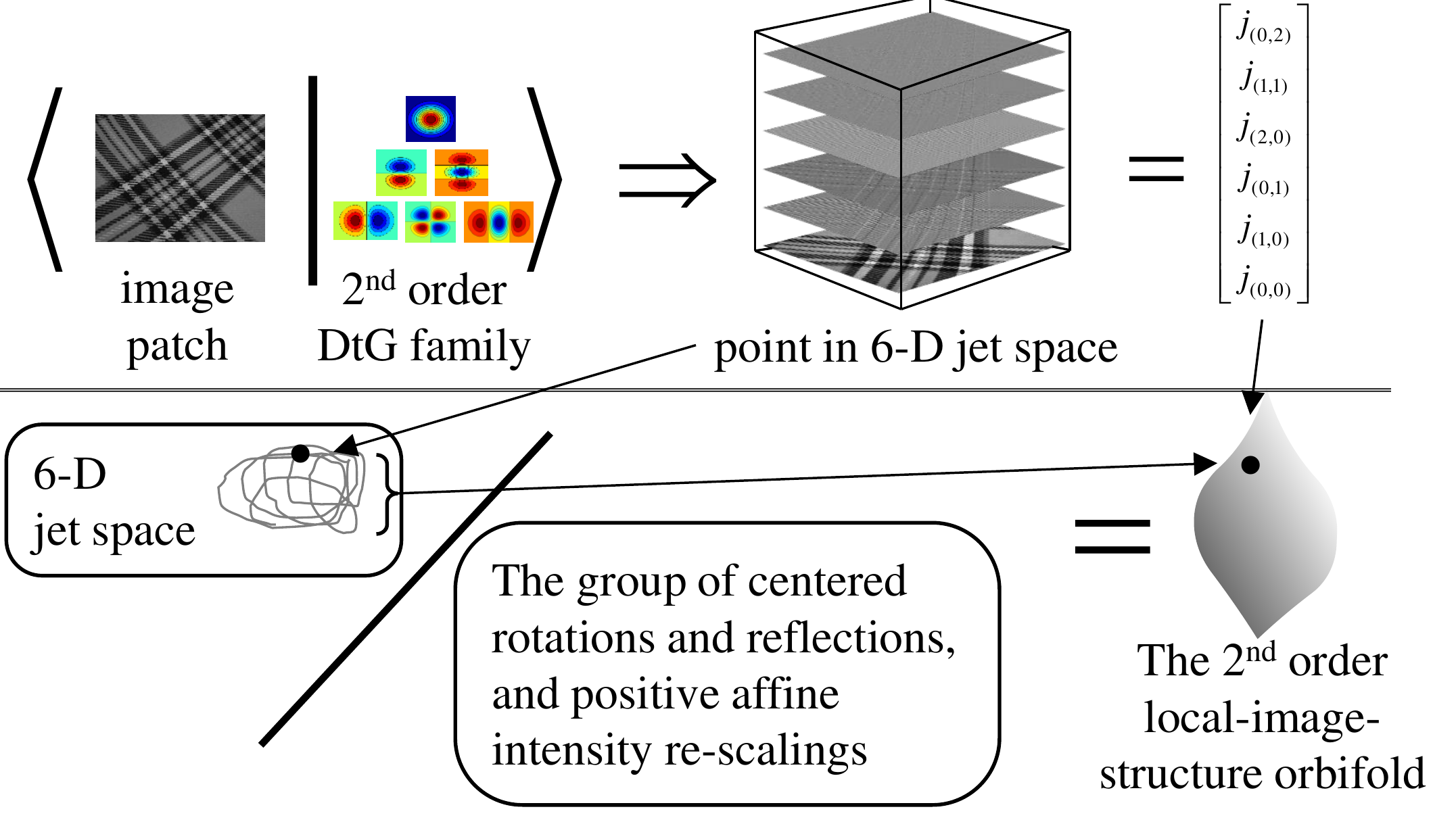}} %trim - left, bottom, right, top
\end{minipage}
\caption{Diagrammatically, (top-left) the measurement of image structure using a bank of DtG filters up to second order and the resulting jet (top-right).}
\label{fig:jet}
\end{figure}

\indent Secondly, the scale space approach to calculate image derivative of a rescaled image can be done alternatively by convolving the original image with the DtGs~($\mathrm{I_{\sigma}^{'} = G_{\sigma}^{'} \ast I}$) as proposed by Young~\cite{young2001gaussian}. He showed that Gaussian derivatives fit more accurately for the measurement of image structure at our receptive fields than other functions such as Gabor function does. The DtGs at scale $\sigma > 0$ is defined in 1-$\mathcal{D}$ by
$G_{\sigma}(x) = G_{\sigma}^{0}(x) = \frac{1}{\sigma\sqrt{2 \pi}} \exp^{\frac{-x^{2}}{2\sigma^{2}}}$.
\begin{equation}
\label{equ:HP}
\begin{aligned}
G_{\sigma}^{m}(x)  = \frac{d^{m}}{dx^{m}}G_{\sigma}(x) 
= (\frac{-1}{\sigma \sqrt{2}})^{m}\mathbf{H}^{m} (\frac{x}{\sigma \sqrt{2}})G_{\sigma}(x)~m \in \mathbb{Z}^{+}
\end{aligned}
\end{equation} \\
\noindent where $G_{\sigma}^{0}(x)$ is the original Gaussian, $m$ is a positive integer, and $H^{m}(x)$ is the $n^{th}$ order Hermite polynomial (Fig.~\ref{fig:Hermite_DtGs}(a))~\cite{martens1990hermite}. It is also useful to perform normalizations of the DtGs, such that $\int{\mid{G_{\sigma}^{k}}(x)\mid}dx$ = 1. In particular, $G_{\sigma}^{0}(x)$ and $G_{\sigma}^{1}(x)$ are the $\ell^{1}$-normalized blurring and differentiating filters, respectively. Hermite transform was originally developed for the mathematical modelling of the properties in early stages of (human) vision~\cite{martens1990hermite}. From the properties of the Hermite polynomials, one sees that the neighborhood functions for even order are \textit{symmetrical}~($G_{\sigma}^{m}(-x) = G_{\sigma}^{m}(x)$), whereas these for odd order are \textit{anti-symmetrical}~($G_{\sigma}^{m}(-x) = -G_{\sigma}^{m}(x)$). For the evaluation of the results of the neighborhood operators (that is, the $G_{\sigma}^{m}(x)$) all over an image, it pays to move to the Fourier domain. The Fourier representation of a convolution with the $n^{th}$ order operator becomes a multiplication with an $n^{th}$ power times a Gaussian frequency envelope, which leads to practical implementations. The DtGs at scale $\sigma > 0$ are defined in 2-$\mathbb{D}$ (Fig.~\ref{fig:Hermite_DtGs}(b)) by
\begin{equation}
\begin{aligned}
& G_{\sigma}^{(m,n)}(x,y) = G_{\sigma}^{m}(x)G_{\sigma}^{n}(y), \hspace{.5cm} m,n \in \mathbb{Z}^{+} \\
&G_{\sigma}(x,y) = G_{\sigma}^{(0,0)}(x,y)
\end{aligned}
\label{equ:2dcov}
\end{equation}
%\begin{figure}[htp]
%\centering
%%\captionsetup{justification=centering}
%\includegraphics[width=.12\textwidth]{Kernel/G00}
%
%(a) $G_{\sigma}^{(0,0)}$ \\
%\medskip
%
%\includegraphics[width=.12\textwidth]{Kernel/G10}\quad
%\includegraphics[width=.12\textwidth]{Kernel/G01}
%
%(b) $G_{\sigma}^{(1,0)}$ \hspace{1.2cm} (c) $G_{\sigma}^{(0,1)}$ \\
%\medskip
%\includegraphics[width=.12\textwidth]{Kernel/G20}\quad
%\includegraphics[width=.12\textwidth]{Kernel/G11}
%\includegraphics[width=.12\textwidth]{Kernel/G02}
%
%(d) $G_{\sigma}^{(2,0)}$ \hspace{.9cm} (e) $G_{\sigma}^{(1,1)}$ \hspace{1cm} (f) $G_{\sigma}^{(0,2)}$ \\
% \caption{(a)-(e) The derivatives of the Gaussian~(DtGs) for 2$\mathbf{D}$ Kernel.}
%\label{fig:DtGs}
%\end{figure}
\noindent Thus, the scale-space method allows to compute the image derivatives of any order at any scale. The convolution formalism should be used when derivatives are required throughout the entire image, whereas an inner product formalism $(\langle .|. \rangle)$ is more acceptable if derivative at a single location is required. For instance, the image derivatives at scale $\sigma$, with respect to the origin is defined as,
\begin{equation}
\begin{aligned}
J_{(m,n)} 
&= (-1)^{(m + n)} \langle G_{\sigma}^{(m,n)} | I \rangle \\
 &= (-1)^{(m + n)} \int_{x,y \in \mathbb{R}} G_{\sigma}^{(m,n)}(x,y) I(x,y)dxdy.
\end{aligned}
\label{equ:Int}
\end{equation}
\noindent Note that the image derivative measurements $J_{m,n}$ (Fig.~\ref{fig:jet}) are dependent on the inner scale $\sigma$, though we do not indicate it with a superscript to prevent cluttered appearance of the equations. Furthermore, we get scale normalized DtGs response ($J_{(m, n)}^{s}$) by multiplying $\sigma^{n + m}$ with the corresponding $J_{(m, n)}$ represented as, 
\begin{equation}
\label{equ:scalejet}
\begin{aligned}
J_{(m, n)}^{s} = \sigma^{n + m} J_{(m, n)}. 
\end{aligned}
\end{equation}

\begin{figure}[htp]
\centering
\captionsetup[subfloat]{farskip=3pt,captionskip=3pt}
\subfloat[]{\includegraphics[width=.16\columnwidth]{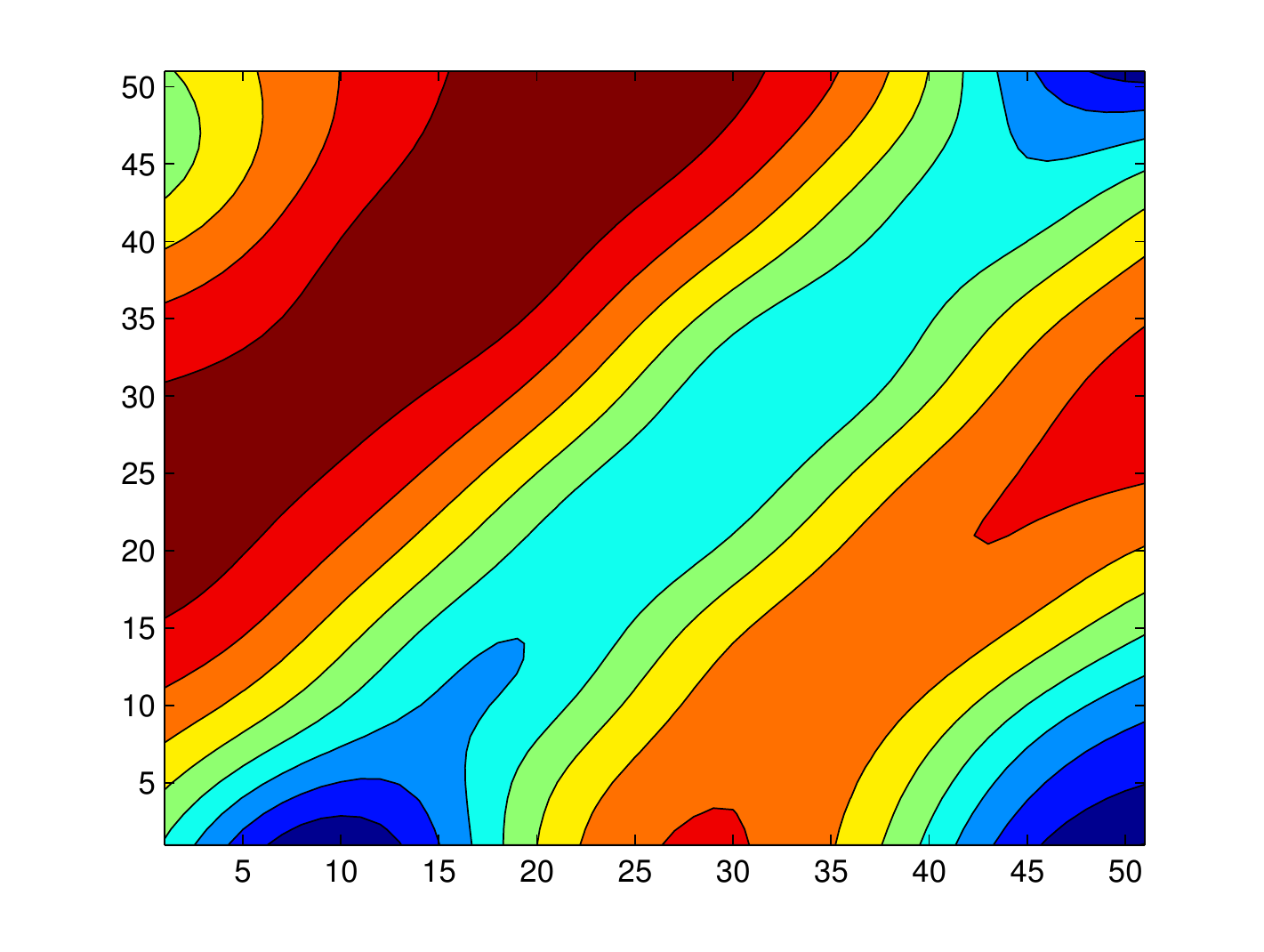}} \hfill
\subfloat[]{\includegraphics[width=.16\columnwidth]{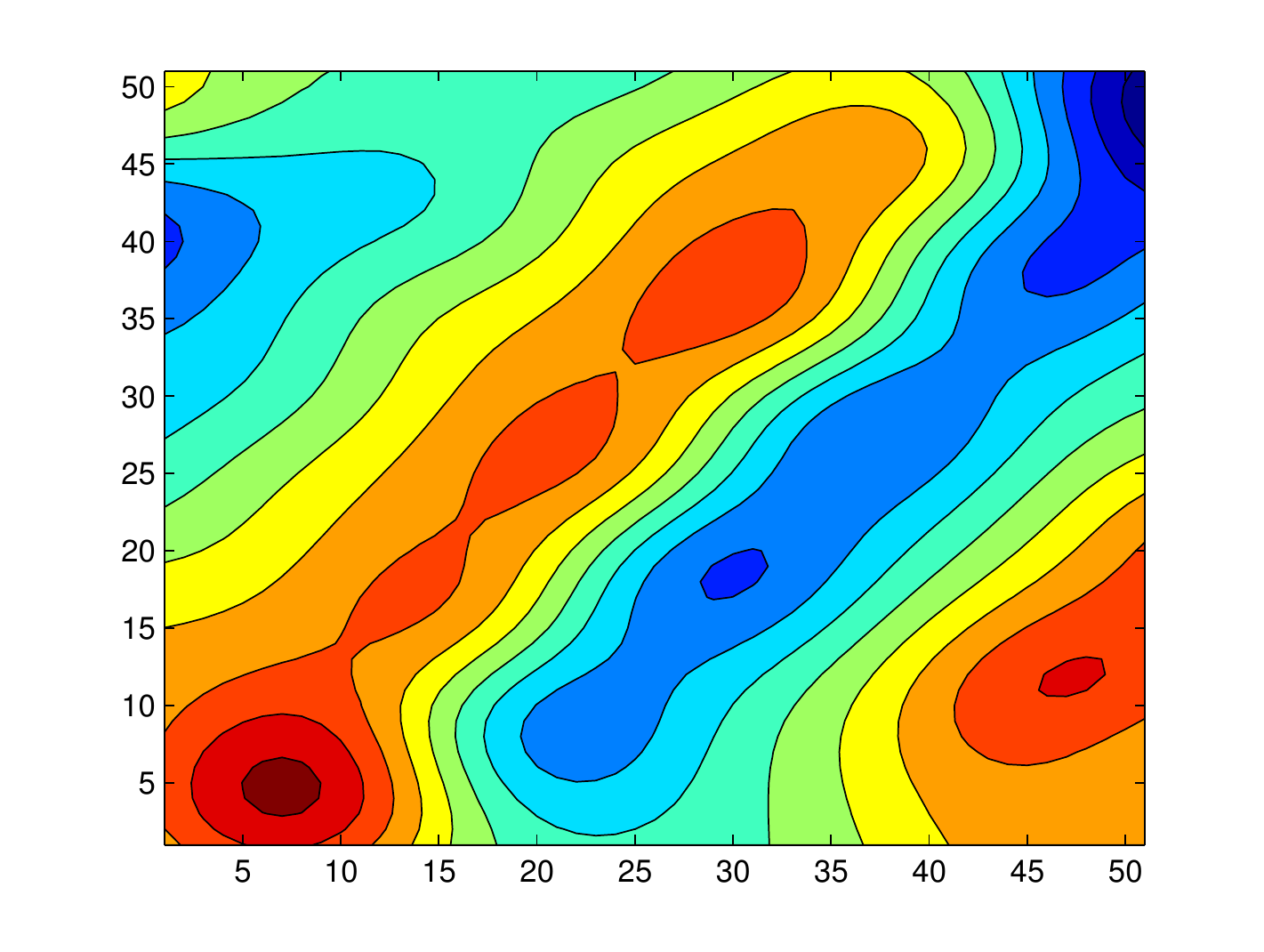}} \hfill
\subfloat[]{\includegraphics[width=.16\columnwidth]{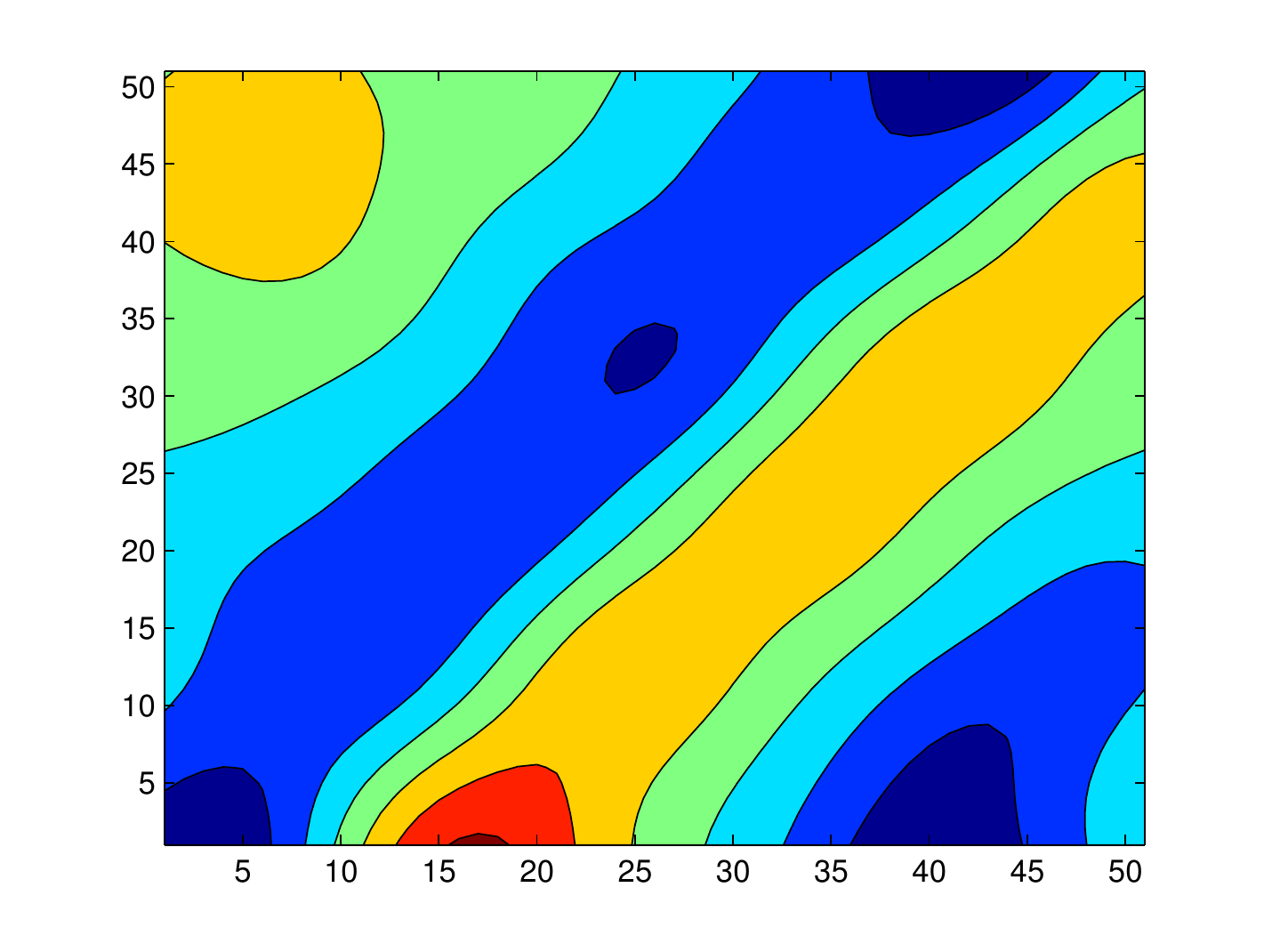}} \hfill
\subfloat[]{\includegraphics[width=.16\columnwidth]{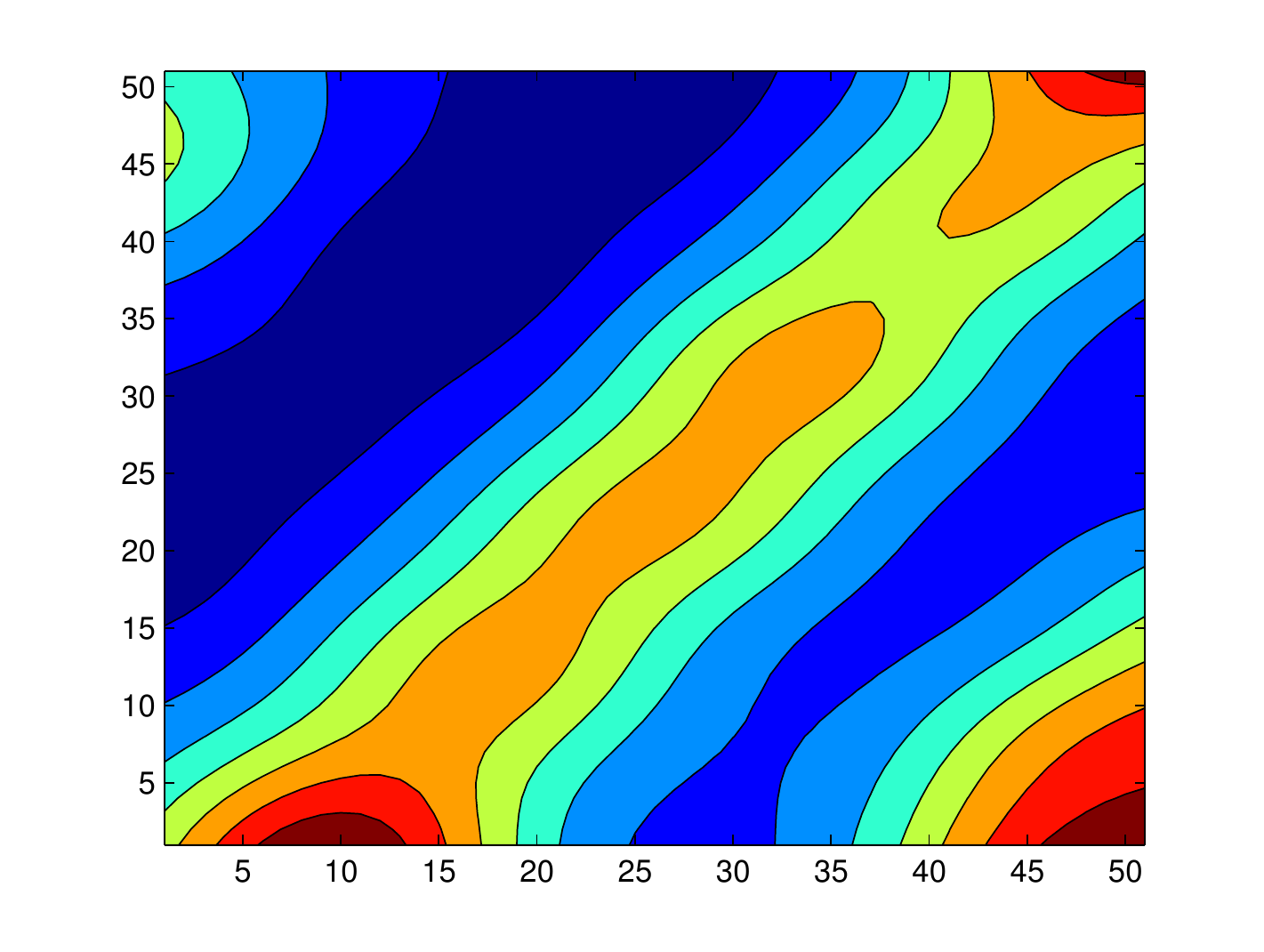}} \hfill
\subfloat[]{\includegraphics[width=.16\columnwidth]{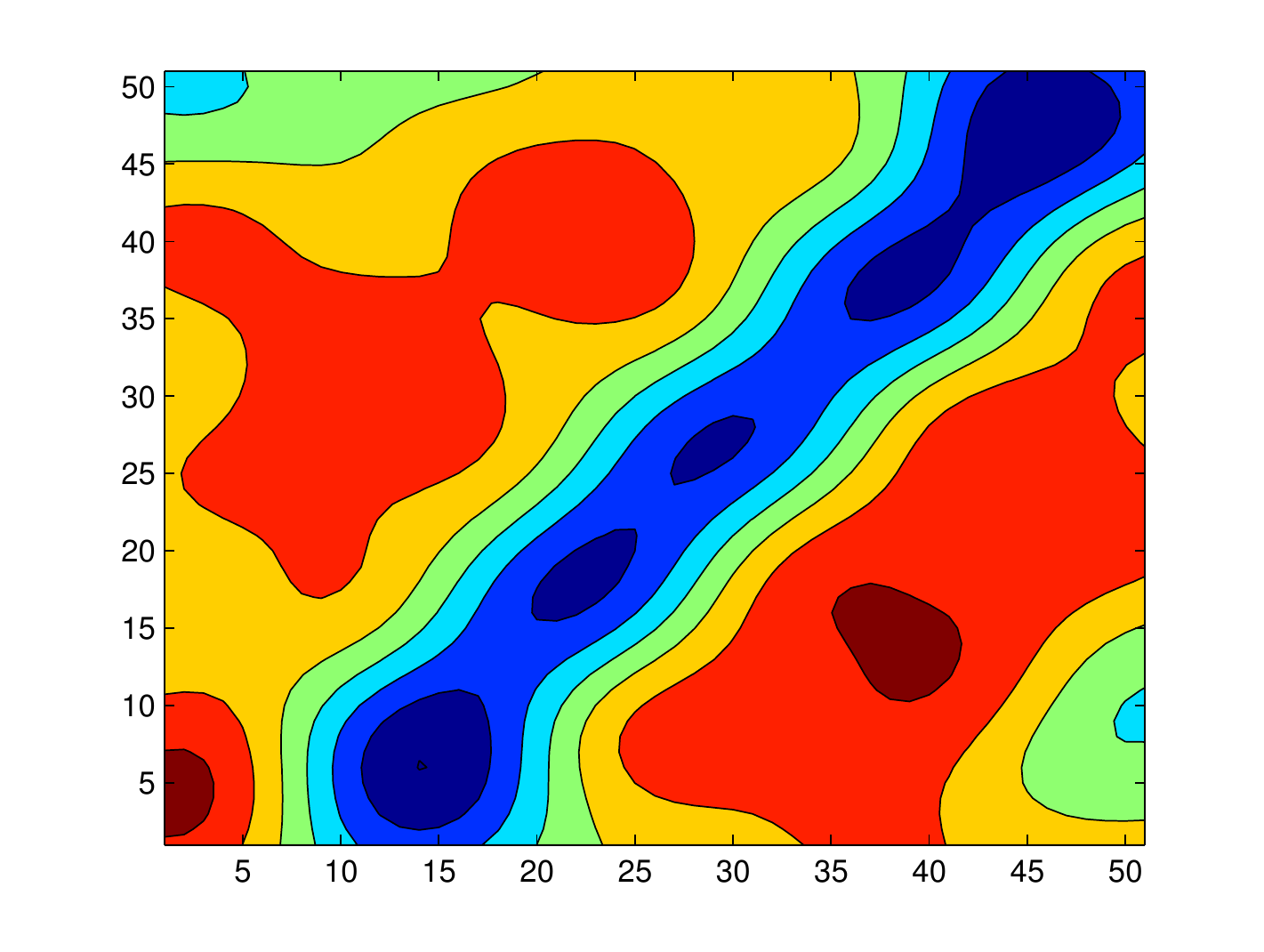}} \hfill
\subfloat[]{\includegraphics[width=.16\columnwidth]{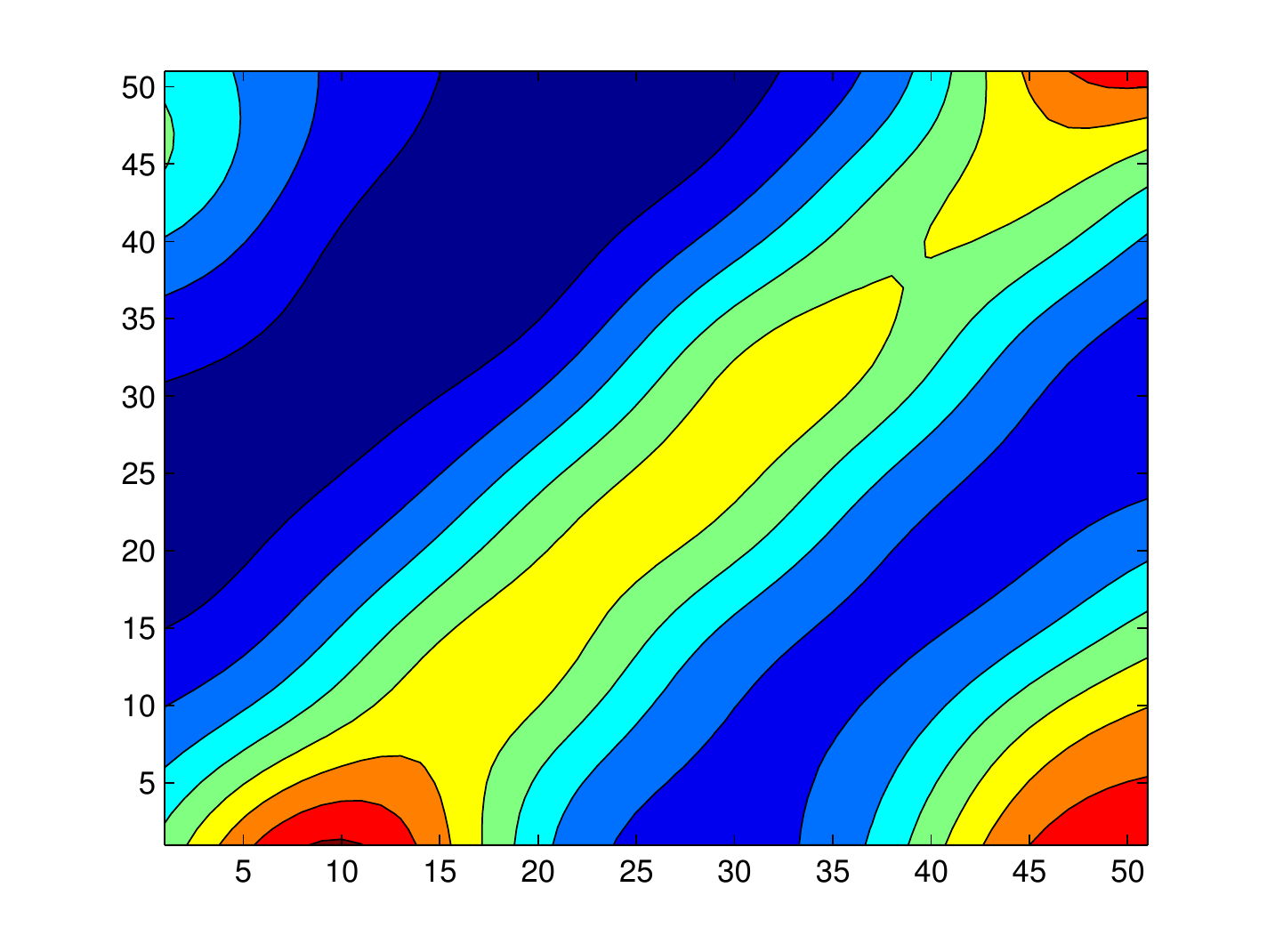}} \\
\caption{(a)-(e). Example of different types of local structure upto $2^{nd}$ order for image patch as shown Fig.~\ref{fig:jet}}.
\label{fig:Cont_plot}
\end{figure}

\noindent It is noted that the DtGs are not orthonormal kernel (e.g, $\langle G_{\sigma}^{(2,0)} | G_{\sigma}^{(0,2)} \rangle = {1}/{(16\pi \sigma^{6})}$) to avoid incorrect presumption. %Typically, one measures with a family of DtG filters with some order. 
For the structure up to $k^{th}$ order, $\{G_{\sigma}^{(m,n)}| 0 \leq m + n \leq k \}$, the vector of DtG responses $J_{(m,n)}^{k} = \langle G_{\sigma}^{m,n}|I\rangle$ is referred to as a local $\mathcal{L}$-jet, where $J_{(m,n)}^{k} \in \mathbb{R}^{\mathcal{L}}, \mathcal{L} = \frac{(k+2)!}{2*k!}$ is called an element of jet space~\cite{florack1996gaussian}. Here, we are only concerned up to $2^{nd}$ order structure, the measurements of which require a set of six DtG kernels denoted as, $$\{ \vec{G}  = (G_{\sigma}^{(0,0)},G_{\sigma}^{(1,0)},G_{\sigma}^{(0,1)}, G_{\sigma}^{(2,0)}, G_{\sigma}^{(1,1)}, G_{\sigma}^{(0,2)}) \}$$, henceforth, called the DtG family (shown in Fig.~\ref{fig:Hermite_DtGs}(b)). The DtG responses calculated from Eqn.~(\ref{equ:scalejet}) are denoted as $\{\vec{J} = (J_{(0,0)}^{s}, J_{(1,0)}^{s}, J_{(0,1)}^{s}, J_{(2,0)}^{s}, J_{(1,1)}^{s}, J_{(0,2)}^{s}) \}$ is called a 6-jet~(shown in Fig.~\ref{fig:jet} (top-left)). We denote $\vec{J} = \langle \vec{G} | I \rangle$ for the jet crop up from the image response $I$. Fig.~\ref{fig:Cont_plot} shows an example of image patches whose local structure is dominated by higher orders (up to $2^{nd}$ order).

\section{Proposed Descriptor}
\label{sec:prop}
\begin{figure*}[htp]
%\captionsetup{justification=left}
\begin{minipage}[b]{1.0\linewidth}
\centering
\centerline{\includegraphics[clip=true, trim=0 50 10 10, width=1.00\linewidth]{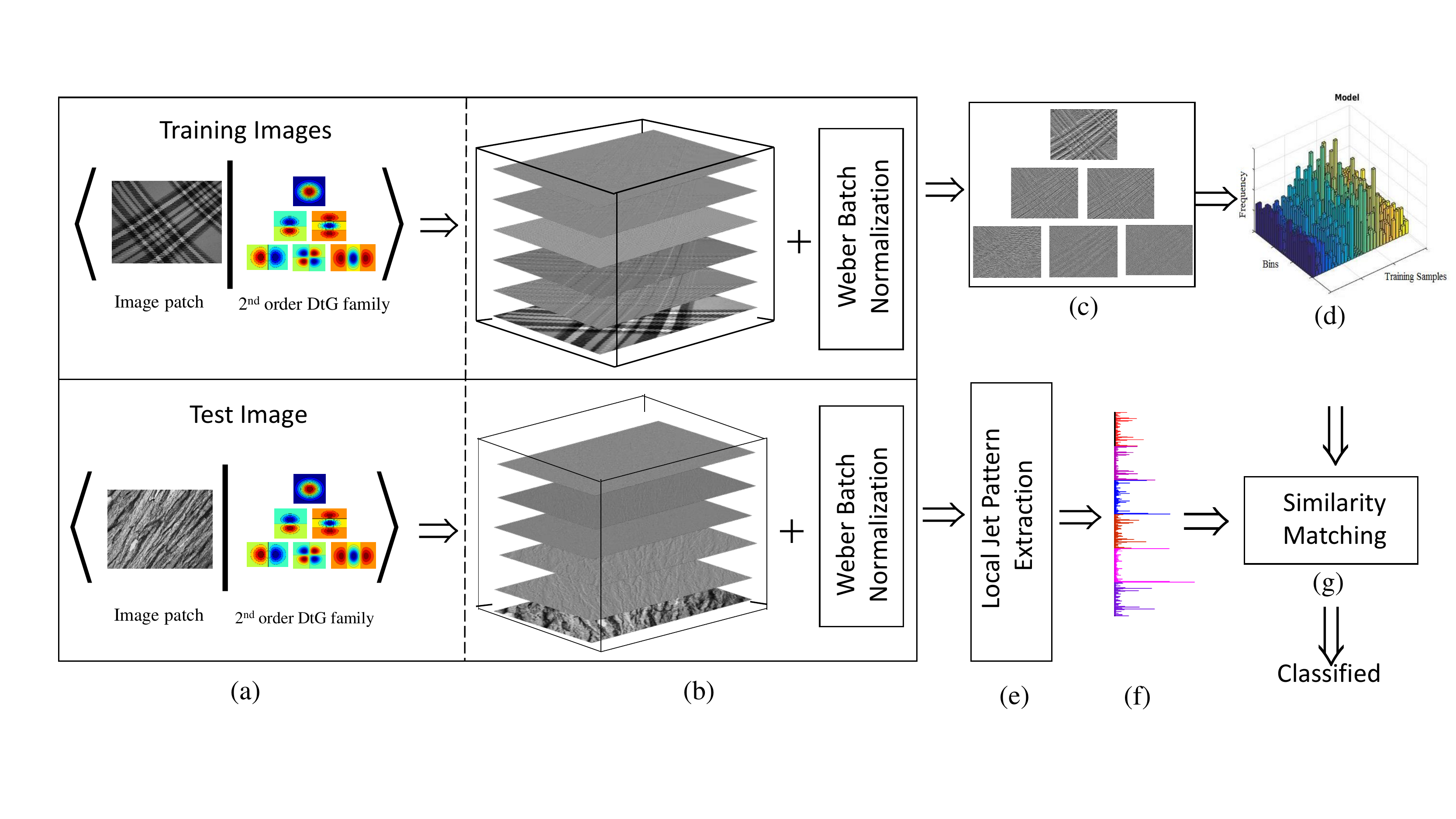}} %trim - left, bottom, right, top
\end{minipage}
\caption{The texture classification framework using proposed local jet pattern(\textsc{Ljp}): (a) texture images with DtGs kernel; (b) the DtGs responses of the texture images (i.e., local jets vector~(\textsc{Ljv})); (c) the local jet pattern~($\textsc{Ljp}_{R,N}^{i,j,l}|l = 1, 2,\ldots, \mathcal{L}$) of individual element of Local Jet Vector~(\textsc{Ljv}) for a training image; (d) the models of frequency distribution for \textsc{Ljp} feature  of all training images; (e) extraction of \textsc{Ljp} for test image; (f) feature distribution of \textsc{Ljp} for test image; (g) similarity matching to classify the test image} 
\label{fig:PLJP}
\end{figure*}

Even the original \textsc{Lbp} scheme is simple and computationally efficient, the $\textsc{Lbp}_{R,N}$ histogram descriptor has the following limitations for an effortless application,

\begin{enumerate}
\item The original $\textsc{Lbp}_{R,N}$ produces a larger histogram, $2^{N}$ different values, even for small neighborhoods, as shown in Table~\ref{Tab:fdim} in order to reduce its' discriminative power whereas increase storage requirements.

\item The original $\textsc{Lbp}_{R,N}$ codes are often not robust to image rotation.

\item $\textsc{Lbp}_{R,N}$ codes are perhaps highly sensitive to noise: the small variation in upper or lower value of the referenced pixel is considered similar to a major contrast.  
\end{enumerate}
\begin{table}[htp]
\caption{Dimensions of Different Descriptors}
% \resizebox{.48\textwidth}{!}{
\begin{tabular}{ c c c c c c c}
 \hline
 Scale & (\textbf{R, N}) & $\textsc{Lbp}_{R,N}$  & $\textsc{Lbp}_{R,N}^{ri}$ & $\textsc{Lbp}_{R,N}^{u2}$ & $\textsc{Lbp}_{R,N}^{riu2}$ & $\textsc{Clbp\_CSM}$\\\hline
1 & (1, 8) & 256  & 36 & 59 & 10 & 200 \\  
2 & (2, 16) & 65536  & 4116 & 243 & 18 & 648\\  
3 & (3, 24) & 16777216 & 699252 & \_ & 26 & 1352\\ 
4 & (4, 32) & $2^{32}$ & large & \_ & 34 & 2312\\
5 & (5, 40) & $2^{40}$ & large & \_ & 42 & 3528\\ \hline 
1-5 &  & infeasible  & infeasible &  & 106 & 8040\\ \hline 
\end{tabular}
% }
\label{Tab:fdim}
\end{table}
\noindent Even the sampling method is similar to that in \textsc{Lbp}, this work proposes a new texture descriptor, called local jet pattern (\textsc{Ljp}) to overcome the aforementioned problems for classification and to address together 1) invariance to scale, translation, and rotation (or reflection), 2) noise tolerance of a texture image. The texture classification framework using proposed \textsc{Ljp} is shown in Fig. \ref{fig:PLJP}. The details of the proposed method are as follows.

\subsection{Local Jet Pattern}
\label{subsec:LJP}
There is no specific scale at which a real-world natural texture should preferably be used. The hidden feature points are potentially detected by searching image feature locally over all scales. Therefore, it is desirable that the image should be represented in a scale space domain, so that a finite class of resolutions can be considered. In this work, we constructed the local jet pattern~(\textsc{Ljp}) to analyze the morphology of a texture image using the spatial derivatives of the corresponding image with different isotropic and anisotropic Gaussian structure. Inspired by the working principle of visual cortex system by Florack \textit{et al.}~\cite{florack1996gaussian}, we use here multi-scale $\mathcal{L}$-jet representation of an image, which is formed using the spatial derivatives. A jet space representation of a texture image is derived from a set of derivative of Gaussian~(DtGs) filter responses up to second order, so called local jet vector~(\textsc{Ljv}), which also satisfies the properties of Scale Space. 
% In other words, the jet can also be interpreted as isolating an image patch with a Gaussian window and then probing it with Hermite function - which is not unlike a windowed Fourier transform. 
To obtain \textsc{Ljp}, initially, we transform the candidate texture image into local $\mathcal{L}$-jet~($\mathcal{L} = 6$) using Eqn.~(\ref{equ:2dcov}), where the element of jet represents the responses of the DtGs~(discussed in Section-\ref{sec:LFJS}) upto $2^{nd}$ order. The scale normalized derivative of 6-jet for a given image $I$ are as follows: $$\vec{J} = (J_{(0,0)}^{s}, J_{(1,0)}^{s}, J_{(0,1)}^{s}, J_{(2,0)}^{s}, J_{(1,1)}^{s}, J_{(0,2)}^{s})$$  represented as a vector $\{\vec{J} = (J^{1}, J^{2},\ldots, J^{\mathcal{L}-1}, J^{\mathcal{L}})\}$ which we refer to as the local jet vector~(\textsc{Ljv}) (shown in Fig.~\ref{fig:PLJP}(b)). The local jet patterns~(\textsc{Ljp}) are computed after a contrast normalization preprocessing step, motivated by Weber's law~\cite{varma2009statistical}. Let $\|J^{(i, j)}\|$ be the $\ell_{2}$ norm of the DtG responses at pixel $(i, j)$. The normalization of the DtGs responses are computed as,
\begin{equation}
\begin{aligned}
\mathbf{J}^{(i,j)} \leftarrow \mathbf{J}^{(i,j)} \times \frac{\log(1 + \frac{L^{(i, j)}}{0.03})}{L^{(i,j)}},
\end{aligned}
\end{equation}
\noindent  where $ L^{(i,j)} = \|\mathbf{J}^{(i, j)}\|_{2} $ represents the DtGs response at pixel $(i, j)$. 
\begin{figure}[htp]
%\captionsetup{justification=left}
\centering
\begin{minipage}[b]{.60\linewidth}
\centerline{\includegraphics[clip = true, trim=20 20 20 20]{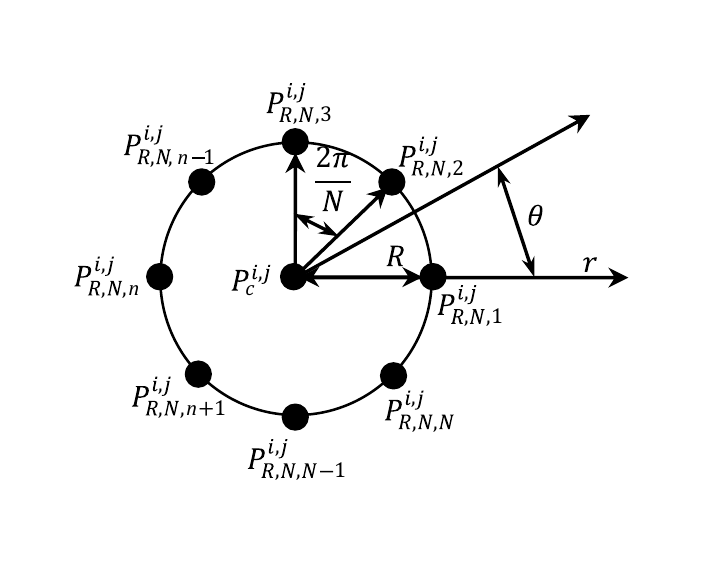}} %trim - left, bottom, right, top
\end{minipage}
\caption{The local neighbors $P_{R,N,n}^{i,j}$ for ($\forall n \in [1, N]$) of a center pixel $P_{c}^{i,j}$ in polar coordinate system.}
\label{fig:lbp}
\end{figure}
Finally, the local jet pattern~(\textsc{Ljp}) for each element in the jet vector for a given center pixel $P_{c}^{i,j}$ at ($i,j$) is computed by comparing its gray value $J_{c}^{i,j}$ with the gray value of a set of $N$ local circularly and equally spaced neighbors $P_{R,N}^{i,j}$ with radius $R$ around the center pixel $P_{c}^{i,j}$ (Fig.~\ref{fig:lbp}). In \textsc{Ljp}, a pixel is considered as a center pixel whose all $N$ local neighbors are present within the image in jet space of dimension $M_x \times M_y$. In Fig.~\ref{fig:lbp}, $P_{R,N,n}^{i,j}$ having gray value $J_{R,N,n}^{i,j}$ represent $n^{th}$ neighbour of $P_{c}^{i,j}$ (i.e. $n^{th}$ component of $P_{R,N}^{i,j}$) where $n$ is a positive integer and $n \in [1, N]$. The spatial coordinate~($x, y$) of $P_{R,N,n}^{i,j}$ with respect to the center, $P_{c}^{i,j}$  is given as, 
\begin{equation}
\begin{aligned}
 x(P_{R,N,n}^{i,j}) = i + r(P_{R,N,n}^{i,j})\times \cos(\theta(P_{R,N,n}^{i,j})) \\
 y(P_{R,N,n}^{i,j}) = j - r(P_{R,N,n}^{i,j})\times \sin(\theta(P_{R,N,n}^{i,j}))
\end{aligned}
\end{equation}
\noindent where $i \in [R + 1, M_{x} - R]$ and $j \in [R + 1, M_{y} - R]$. $r(P_{R,N,n}^{i,j})$ and $\theta(P_{R,N,n}^{i,j})$ denote the polar coordinates of $P_{R,N,n}^{i,j}$ where $n= 1, 2, \ldots, N$ and are computed as
\begin{equation}
\begin{aligned}
r(P_{R,N,n}^{i,j}) = R \\
\theta(P_{R,N,n}^{i,j}) = (n - 1) \times \frac{2\pi}{N} 
\end{aligned}
\end{equation}
Let $J_{c}^{i,j,l}$ be a center pixel $(i,j)$ of $l^{th}$ element of jet vector~($l\in [1,\mathcal{L}]$) and $J_{R,N,n}^{i,j,l}$ be the $n^{th}$ neighbor of $J_{c}^{i,j,l}$ corresponding to the $N$ sampling points with radius $R$. The array of $N$ neighbors are defined as, 
$$ \bar{J}_{R,N}^{i,j,l} = [J_{R,N,1}^{i,j,l},J_{R,N,2}^{i,j,l},\ldots, J_{R,N,N}^{i,j,l}]$$
The \textsc{Ljp} response for a given center pixel $J_{c}^{i,j,l}$ of $l^{th}$ element of local jet vector~(\textsc{Ljv}) is calculated as,
\begin{equation}
\begin{aligned}
\textsc{Ljp}_{R,N}^{i,j,l} = \sum_{n = 1}^{N}2^{n - 1} \times sign(J_{R,N,n}^{i,j,l}- J_{c}^{i,j,l})
\end{aligned}
\end{equation}
where $sign$ is a \textit{unit step} function to denote whether a given input is positive or not, and is defined as,

$$sign(z) = \begin{cases} 1, &  z \geq 0\\ 0, & z < 0 \end{cases}$$

\noindent Note that the range of $\textsc{Ljp}$ depends on the number of neighboring sampling points~($N$) around the center $J_{R,N}^{i,j,l}$ and radius $R$ to form the pattern and their values lie between $0$ to $2^{N - 1}$. In other words, the range of \textsc{Ljp} is [$0, 2^{N-1}$]. We compute the \textsc{Ljp} for all elements of \textsc{Ljv}. Fig.~\ref{fig:PLJP}(c) shows the computed local jet patterns~(i.e. $\textsc{Ljp}_{R,N}^{i,j,l}\vert l = 1, 2, \ldots, \mathcal{L}$) from each element of \textsc{Ljv} (Fig.~\ref{fig:PLJP}(b)) for a candidate training texture image (Fig.~\ref{fig:PLJP}(a)).

\subsection{\textsc{Ljp} Feature Vector}
\label{subsec:LJPFV}
To reduce the classification computation cost using local jet pattern~(\textsc{Ljp}) of a texture image, we need to compute the frequency of local jet pattern~(\textsc{Ljp}) for each element of local jet vector~(\textsc{Ljv})~($\bar{J} = (J^1, J^2,\ldots,J^{\mathcal{L}-1},J^\mathcal{L})$). The frequency of local jet pattern~(\textsc{Ljp}) ($H$) of $2^{N}$ dimension is calculated for every pixel of $\textsc{Ljp}$ image $I$ of size $(M_x \times M_y)$. The normalized $\textsc{Ljp}$ histogram feature vector ($\textbf{H}_{R,N}^l$) for $l^{th} \in [1, \mathcal{L}]$ element of \textsc{Ljv} (i.e. $\textsc{Ljp}_{R,N}^{i,j,l}$), is computed using the following equation,
\begin{equation}
H_{R,N}^{l}(\Omega) = \sum_{i = R+1}^{M_{x}-R}\sum_{j = R+1}^{M_{y}-R} f(\textsc{Ljp}_{R,N}^{i,j,l}(i ,j),\Omega)
\label{equ:ljp_hist}
\end{equation}

\noindent where $\forall {\Omega} \in [0, 2^{N} - 1]$ and $f(u,v)$ is a function given by
$$f(u,v) = \begin{cases} 1, & if~u == v\\ 0, & otherwise. \end{cases}$$
\indent The final \textsc{Ljp} feature vector is constructed by concatenating all $H_{R,N}^{l}$ where $l = 1, 2,\ldots, \mathcal{L}$. Since the number of bins in Eqn.~(\ref{equ:ljp_hist}) of $\textsc{Ljp}_{R,N}^{i,j,l}$ for each element of \textsc{Ljv}~($l \in [1, \mathcal{L}]$) is 256 when R = 1, N = 8. Therefore, the dimension of the \textsc{Ljp} feature vector becomes high i.e. 256 $\times$ 6 = 1536. To cut down the number of bins, we adopted \textit{uniform} pattern scheme \cite{ojala2002multiresolution}, which gives 59 bins for each element of \textsc{Ljv} and the final normalized \textsc{Ljp} feature vector dimension is reduced to 59 $\times$ 6 = 354. Since the zeroth order terms of DtGs response does not carry any geometrical information, those are discarded. Therefore, the feature vector dimension is further reduced to 59 $\times$ 5 = 295. 
\begin{figure}[htp]
\centering
\captionsetup[subfloat]{farskip=3pt,captionskip=3pt}
\subfloat[Sample1]{\includegraphics[width=0.20\linewidth]{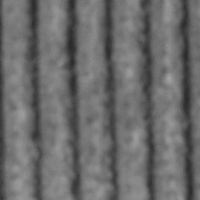}%
\label{fig:test_img1}}
\hspace{1cm}
\subfloat[Sample2]{\includegraphics[width=0.20\linewidth]{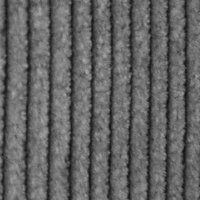}%
\label{fig:test_img2}}
\hspace{1cm}
\subfloat[Sample3]{\includegraphics[width=0.20\linewidth]{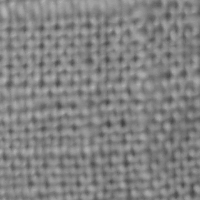}%
\label{fig:test_img3}}
\hfill
\subfloat[]{\includegraphics[width=0.49\linewidth]{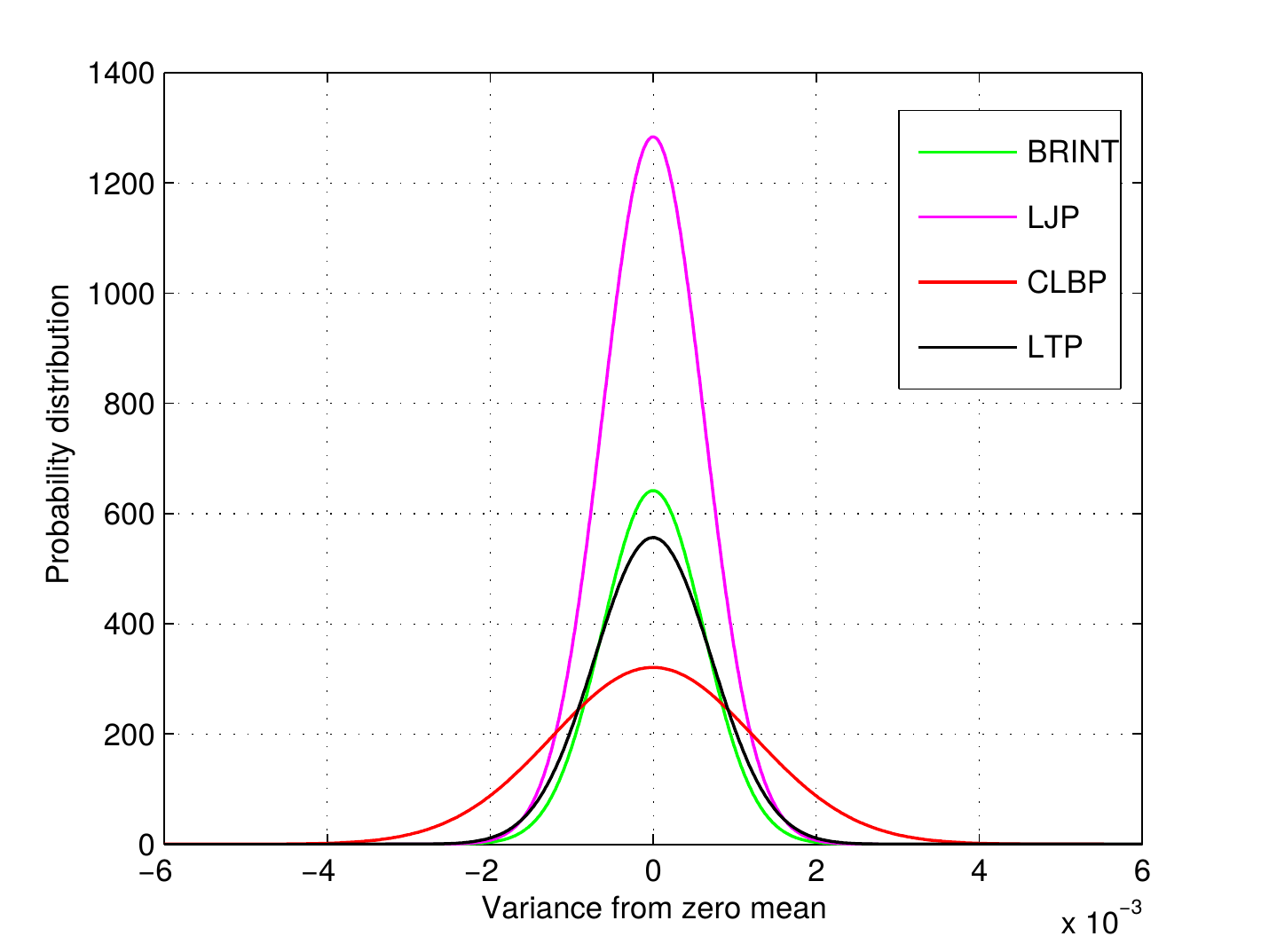}%
\label{fig:intra_class}}
\hfill
\subfloat[]{\includegraphics[width=0.49\linewidth]{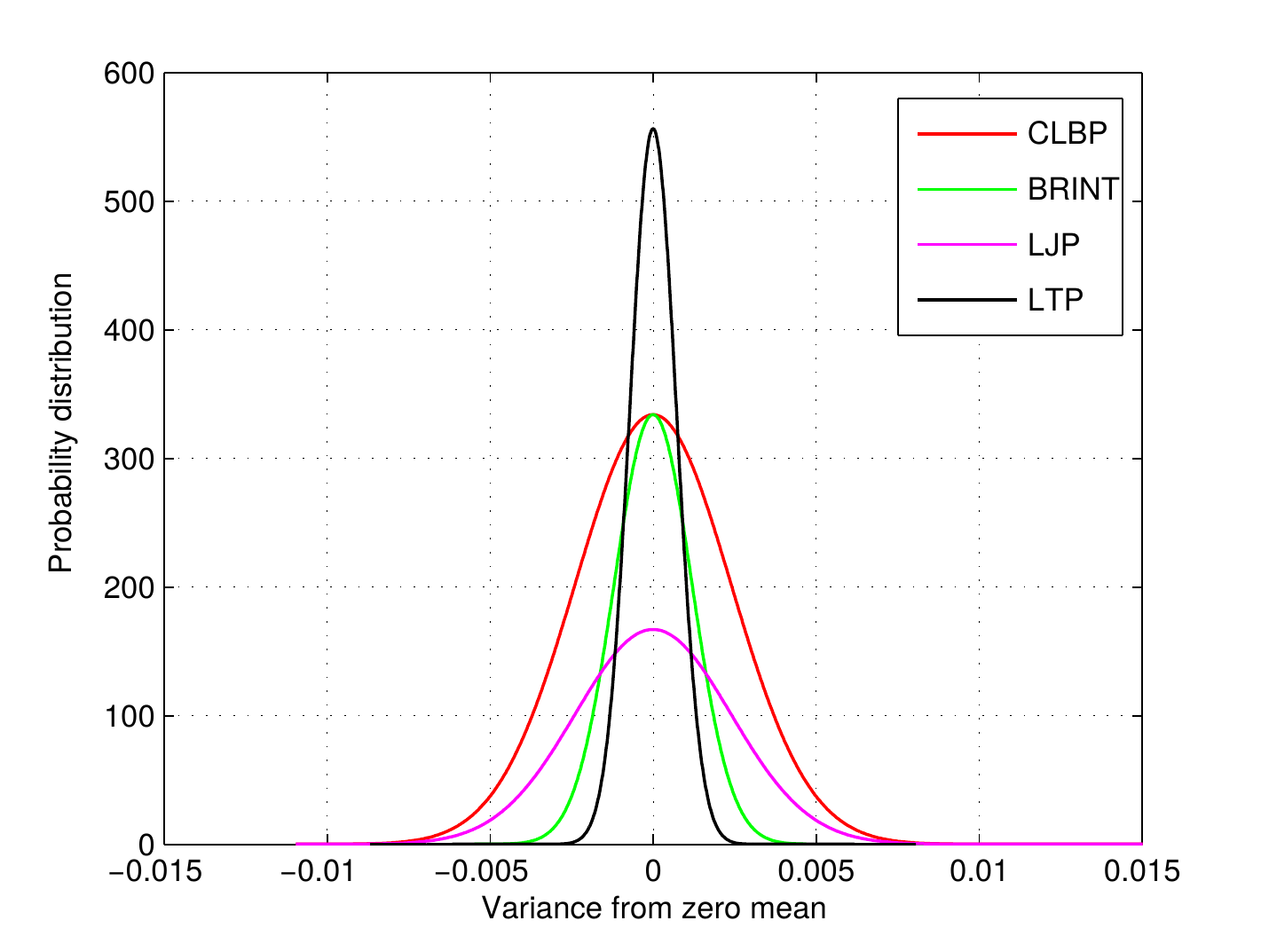}%
\label{fig:inter_class}}
\hfill
\caption{Illustration of the behavior of \textsc{Ltp}, \textsc{Brint}, \textsc{Clbp}, and \textsc{Ljp} for inter and intra class sample images; (a)-(c) Three texture images taken from KTH-TIPS~\cite{hayman2004significance} texture database where Sample1 and Sample2 belong to the same class whereas Sample1 and Sample3 belong to the different classes; (d)-(e) represents the probability distributions of feature vectors difference with respect to zero mean of different methods for intra and inter class images respectively.}
\label{fig:KTH_sample}
\end{figure}

In Appendix.~\ref{subsec:EST}, we observe that $J_{(0,0)}$ and $J_{(1,1)}$ of \textsc{Ljv} are invariant to rotation, or reflec individually. Now,  considering \textsc{Ljp} for all the elements of \textsc{Ljv}, i.e. $\mathcal{L}$-jet ($\mathcal{L}$ = 6) together, is invariant under scale, rotation, or reflation of the image surface. Therefore, the final \textsc{Ljp} feature vector is also invariant to scale, rotation, or reflation of the texture image.    

We computed and compared the probability distributions of feature vector difference of proposed and state-of-the-art features using sample images of intra and inter class and shown in Fig.~\ref{fig:KTH_sample}. The horizontal axis indicates the deviation from zero mean and the vertical axis indicates the probability of feature vector difference for one sample image to another by a particular amount of deviations. The probability at zero mean with high amplitude represents higher similarity between feature vectors whereas larger deviation from zero mean implies the less similarity. Fig.~\ref{fig:KTH_sample} shows that \textsc{Ljp} is discriminative as it perfectly differentiates inter class variations (Sample1 and Sample3) and at the same time better matches for intra class images (Sample1 and Sample2).

Assuming image of size $M_{x} \times M_{y}$ and the kernel of size $P \times Q$, the algorithmic complexity of the proposed descriptors is computed as follows. To create the $\mathcal{L}$-jet representation of input image in step 4 of Algorithm~\ref{algo:ljp} requires roughly $\mathcal{L}M_{x}M_{y}PQ$ multiplications and additions. As 2-$\mathbb{D}$ DtGs kernel is separable, filtering is done in two steps. The first step requires about $\mathcal{L}M_{x}M_{y}P$ multiplications and additions while the second requires about $\mathcal{L}M_{x}M_{y}Q$ multiplications and additions, making a total of $\mathcal{L}M_{x}M_{y}(P + Q)$. Since $\mathcal{L}$ and kernel size are constant, the complexity becomes $O(M_{x}M_{y})$. The creation of \textsc{Ljp} pattern, as presented in step 11 of Algorithm~\ref{algo:ljp}, requires roughly $\mathcal{L}NM_{x}M_{y}$  multiplications and additions, and the complexity becomes $O(M_{x}M_{y})$, where $N$ is the number of evenly spaced circular neighbor pixels to each center. As step 13 of Algorithm~\ref{algo:ljp} takes $O(M_{x}M_{y})$ to create histogram of \textsc{Ljp}, the time complexity of proposed \textsc{Ljp} descriptor is $3M_{x}M_{y} \approx O(M_{x}M_{y})$.

\begin{algorithm}[htp]
\SetAlgoLined
\KwData{The texture image ($I$) }
\KwResult{Normalized Local Jet Pattern~(\textsc{Hljp}) Descriptor }
 initialization~(R = 1, N = 8, k = 2, m = 0, n = 0)\;
 
 \While{(m $\leq$ 2 and n $\leq$ 2 and (m + n) $\leq$ 2 )}{
  {
   Create DtGs Kernel ($G_{\sigma}^{m, n}$) upto k = $2^{nd}$ Order\;
   
   $J_{(m, n)} = (-1)^{m + n}\langle 		      G_{\sigma}^{m, n}| I \rangle$\;
  }
 } 
$J_{(m, n)}^{s} = {\sigma}^{m + n}J_{(m, n)}$ 
 
%Contrast Normalization~$\mathbf{J}^{(i,j)} \leftarrow \mathbf{J}^{(i,j)} \times \frac{\log(1 + \frac{L^{(i, j)}}{0.03})}{L^{(i,j)}}$\;
initialization~(i = R+1, j = R+1, $l$ = 1, $\mathcal{L}$ = $\frac{(k+2)!}{2 \times k!}$)\;	
\While{(l $\leq$ $\mathcal{L}$)}{
  {
  \While{(i $\leq$ $M_x - R$ and j $\leq$ $M_y - R$)}{
  $\mathbf{J}^{(i,j,l)} \leftarrow \mathbf{J}^{(i,j,l)} \times \frac{\log(1 + \frac{L^{(i, j)}}{0.03})}{L^{(i,j)}}$\;
  $\textsc{Ljp}_{R, N}^{(i, j, l)} = \sum_{n = 1}^{N}2^{(n - 1)} \times sign(J_{R,N,n}^{i,j,l}- J_{c}^{i,j,l})$\;
  } 
  $\mathrm{H_{R,N}^{l}}$ = $\textsc{Hist}\mathrm{(\textsc{Ljp}_{R, N}^{i, j, l})}$\;
  \textsc{Hljp} = $\textit{concateNormalize\textsc{Ljp}}(H_{R,N}^{l})$\;
  }
 }

 \caption{Algorithm to extract $\textsc{Ljp}_{R,N}^{i,j,l}$ descriptor, where N is the number of circular samples taken over circle of radius R.}
 \label{algo:ljp}
\end{algorithm}

\subsection{Comparing Distribution of local Descriptor}
\label{sec:FM}
To compare distributions of local descriptors, we need to find the distributions of \textsc{Ljp}s for training (model) (Fig.~\ref{fig:PLJP}(d)) and test (sample) images (Fig.~\ref{fig:PLJP}(f)).
% After computing the \textsc{Ljp} descriptors as elaborated in the previous section, we need to represent their distributions in the training (model) (Fig.~\ref{fig:PLJP}(d)) and test (sample) images (Fig.~\ref{fig:PLJP}(f)). 
A non-parametric statistical test can measure the goodness-of-fit using the dissimilarity of sample and model histograms (Fig.~\ref{fig:PLJP}(g)). To evaluate the fit between two histograms, there are several metrics such as log-likelihood ratio, histogram intersection, and chi-square~($\chi^2$) statistic~\cite{ojala2002multiresolution}. In this work, two non-parametric classifiers is used to analyze the actual classification performance. First, the nearest neighbor classifier (\textsc{Nnc}) with the chi-square~($\chi^2$) distance~\cite{guo2010completed} is utilized to evaluate the effectiveness of proposed feature. The comparative distributions of two samples $H_{x} = {x_1, \ldots, x_N}$ and $H_{y} = {y_1, \ldots, y_N}$ using  the $\chi^2$ distance is defined as,
\begin{equation}
\label{equ:DM}
\begin{aligned}
D(H_{x}, H_{y}) = \sum_{i = 1}^{N} \frac{(x_{i} - y_{i})^2}{x_{i} + y_{i}}
\end{aligned}
\end{equation}
\noindent where $N$ represents the number of bins, $H_x$ and $H_y$ represent the extracted features of a model and a test sample, respectively. The class of test sample $H_y$ is labeled to that class of model that makes smallest $\chi^2$-distance. 

Then another advanced classifier, called nearest subspace classifier \textsc{Nsc} \cite{wright2009robust}, is implemented to achieve an improved classification performance. In order to avoid the over emphasizing patterns with large frequency, the proposed feature is preprocessed before fitting to \textsc{Nsc}, similar to that in~\cite{guo2016robust}:
\begin{equation}
\overline{X_{k}} = \sqrt{X_{k}},~ k = 1, 2,...,N
\end{equation}
\noindent where $N$ and $X_{k}$ are the number of bins and the original frequency of the $\textsc{Ljp}$ at $k^{th}$ bin, respectively. At first the nearest subspace classifier~(\textsc{Nsc}) computes the distance of test sample, y to the $c^{th}$ class and calculates the projection residual $r_{c}$ from y to the orthogonal principle subspace $B_{c} \in \mathbb{R}^{N \times n}$ of the training sets $X_c$, which is spanned by the principal eigenvectors of $\sum_{c} = X_cX_c^{T}$ for the $c^{th}$ class, given as follows,
\begin{equation}
\mathrm{ r_c = \|(I - P_{B_{c}})y\|_{2} = \|(I - B_cB_c^{T})y\|_{2}}
\end{equation}

\noindent where $P = I \in \mathbb{R}^{N \times N}$ is a identity matrix where $N$ rows are selected uniformly at random. The test sample $y$ is then assigned to the one of the $\mathbf{C}$ classes with the smallest residual among all classes, i.e.
\begin{equation}
i^{*} = arg\min_{c = 1, \ldots, C} r_c
\end{equation}

\section{Experimental Evaluation}
\label{sec:results}

\subsection{Texture Databases}
\label{subsec:texture_db}
To evaluate the classification power of the proposed descriptor, a series of experiments is carried out on five large and commonly used texture databases: Outex\_TC-00010~(TC10) \cite{ojala2002outex}, Outex\_TC-00012~(TC12) \cite{ojala2002outex}, Brodatz album \cite{brodatz1966textures}, KTH-TIPS \cite{hayman2004significance}, and CUReT \cite{dana1999reflectance} texture databases. The Brodatz database is a well known challenging database to test texture classification algorithms because it contains comparatively large number of texture classes, having the lack of intra-class variation and small number of samples per class. The performance of different methods are evaluated in term of classification accuracy using K-fold cross-validation test along with two non-parametric classifiers, \textsc{Nnc} with Chi-Squre ($\chi^2$) distance and \textsc{Nsc}. In $k$-fold cross-validation test, the feature set is randomly partitioned into $k$ equal sized subsets ($k=10$). Out of the $k$ subsets, a single subset is retained as the validation data for testing the classifier, and the remaining $(k - 1)$ subsets are used as training data. The average of the classification accuracies over $k$ rounds give us a final cross-validation accuracy. The K-fold cross-validation process provides a more accurate picture of the classification performance. We have normalized each input image to have an average intensity of 128 and a standard deviation of 20 \cite{ojala2002multiresolution}. In VZ-MR8 and VZ-Patch methods, the texture samples are normalized to have an average intensity of 0 and a standard deviation of 1~\cite{varma2005statistical, varma2007locally, varma2009statistical}. This is done to remove global intensity and contrast. The texture classification accuracies of the proposed descriptor are compared with $\textsc{Lbp}_{R,N}$~\cite{ojala2002multiresolution}, $\textsc{Lbp}_{R,N}^{u2}$, $\textsc{Dlbp}_{R,N}$ \cite{liao2009dominant}, $\textsc{Lbp}_{R,N}^{sri\_su2}$~\cite{li2012scale}, multiscale $\textsc{Clbp}\_S_{R,N}^{riu2}/M_{R,N}^{riu2}/C(1,8 + 3, 16 + 5, 24)$~\cite{guo2010completed} and other state-of-the-art methods. The details of experimental setups are given as follows: 
\begin{table}[htp]
\centering
\caption{Summary of Texture Database used in Experiment \#1}
\label{tab:exp1}
\resizebox{\columnwidth}{!}{
\begin{tabular}{|c|c|c|c|c|c|c|c|}
\hline
\begin{tabular}[c]{@{}c@{}}Texture\\ Database\end{tabular} & \begin{tabular}[c]{@{}c@{}}Image\\ Rotation\end{tabular} & \begin{tabular}[c]{@{}c@{}}Illumination\\ Variation\end{tabular} & \begin{tabular}[c]{@{}c@{}}Scale \\ Variation\end{tabular} & \begin{tabular}[c]{@{}c@{}}Texture\\ Classes\end{tabular} & \begin{tabular}[c]{@{}c@{}}Sample \\ Size (pixels)\end{tabular} & \begin{tabular}[c]{@{}c@{}}Samples\\ per Class\end{tabular} & \begin{tabular}[c]{@{}c@{}}Total\\ Samples\end{tabular} \\ \hline
Outex\_TC10                                                & \checkmark                                               &                                                                  &                                                            & 24                                                        & 128 x 128                                                       & 180                                                         & 4320                                                    \\ \hline
Outex\_TC12                                                & \checkmark                                               & \checkmark                                                       &                                                            & 24                                                        & 128 x 128                                                       & 200                                                         & 4800                                                    \\ \hline
\end{tabular}
}
\end{table}

\noindent \textsc{\textbf{Experiment}} \#1: There are 24 distinct homogeneous texture classes selected from the Outex texture databases~\cite{ojala2002outex}, each image having the size of 128$\times$128 pixels. \textbf{Outex\_TC\_0-0010 (Outex\_TC10)} contains texture images under illuminant ``inca'' whereas \textbf{Outex\_TC\_00012 (Outex\_TC12)} contains texture images with 3 different illuminants~(``inca'', ``horizon'', and ``t184''). Both of the Outex test suit images are collected under 9 different rotation angels ($0^{\degree}$, $5^{\degree}$, $10^{\degree}$, $15^{\degree}$, $30^{\degree}$, $45^{\degree}$, $60^{\degree}$, $75^{\degree}$, and $90^{\degree}$) in each texture class. The test suites \textbf{Outex\_TC\_00010 (Outex\_TC10)}, and  \textbf{Outex\_TC\_00012 (Outex\_TC12)} are summarized in Table~\ref{tab:exp1}. 

\begin{figure}[htp]
\centering
 \includegraphics[width=.10\columnwidth]{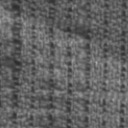}
 \includegraphics[width=.10\columnwidth]{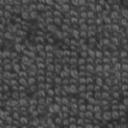} 
 \includegraphics[width=.10\columnwidth]{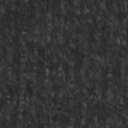} 
 \includegraphics[width=.10\columnwidth]{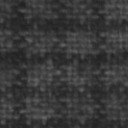}   
 \includegraphics[width=.10\columnwidth]{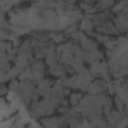} 
 \includegraphics[width=.10\columnwidth]{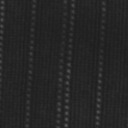}
 \includegraphics[width=.10\columnwidth]{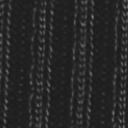} 
 \includegraphics[width=.10\columnwidth]{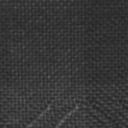} 
 \\ \medskip 
\includegraphics[width=.10\columnwidth]{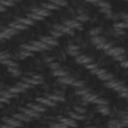} 
\includegraphics[width=.10\columnwidth]{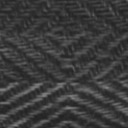} 
\includegraphics[width=.10\columnwidth]{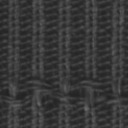}            \includegraphics[width=.10\columnwidth]{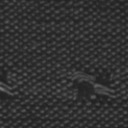} 	 
\includegraphics[width=.10\columnwidth]{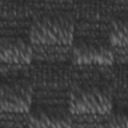} 
\includegraphics[width=.10\columnwidth]{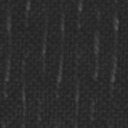} 
\includegraphics[width=.10\columnwidth]{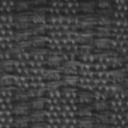}    \includegraphics[width=.10\columnwidth]{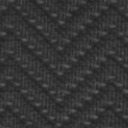} 
\\  \medskip
\includegraphics[width=.10\columnwidth]{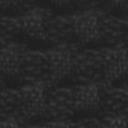} 
\includegraphics[width=.10\columnwidth]{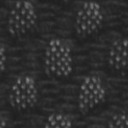} 
\includegraphics[width=.10\columnwidth]{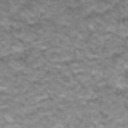}    \includegraphics[width=.10\columnwidth]{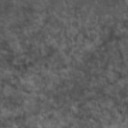} 
\includegraphics[width=.10\columnwidth]{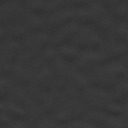} 
\includegraphics[width=.10\columnwidth]{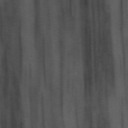} 
\includegraphics[width=.10\columnwidth]{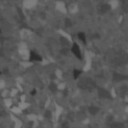}    \includegraphics[width=.10\columnwidth]{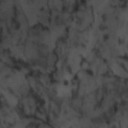} 
\caption{24 texture images randomly taken from each class of Outex\_TC10 and Outex\_TC12 database}
\label{fig:TC_images}
\end{figure}

\noindent \textsc{\textbf{Experiment}} \#2: \textbf{Brodatz} \cite{brodatz1966textures} album contains 32 homogeneous texture classes where 
%\footnote{The 32 Brodatz textures are Bark, Beachsand, Beans, Burlap, D10, D11, D4, D51, D52, D5, D6, D95, Fieldstone, Grass, Ice, Image09, Image15, Image17, Image19, Paper, Peb54, Pigskin, Pressedcl, Raffia2, Raffia, Reptile, Ricepaper, Seafan, Straw2, Tree, Water, Woodgrain;}.
each image is partitioned into 25 non-overlapping sub-images of size $128 \times 128$, and each sub-image is down-sampled to $64 \times 64$ pixels. 

\begin{table}[htp]
\centering
\caption{Summary of Texture Database used in Experiment \#2}
\label{tab:exp2}
\resizebox{\columnwidth}{!}{
\begin{tabular}{|c|c|c|c|c|c|c|c|}
\hline
\begin{tabular}[c]{@{}c@{}}Texture\\ Database\end{tabular} & \begin{tabular}[c]{@{}c@{}}Image\\ Rotation\end{tabular} & \begin{tabular}[c]{@{}c@{}}Illumination\\ Variation\end{tabular} & \begin{tabular}[c]{@{}c@{}}Scale \\ Variation\end{tabular} & \begin{tabular}[c]{@{}c@{}}Texture\\ Classes\end{tabular} & \begin{tabular}[c]{@{}c@{}}Sample \\ Size (pixels)\end{tabular} & \begin{tabular}[c]{@{}c@{}}Samples\\ per Class\end{tabular} & \begin{tabular}[c]{@{}c@{}}Total\\ Samples\end{tabular} \\ \hline
KTH-TIPS                                                   & \checkmark                                               & \checkmark                                                       & \checkmark                                                 & 10                                                        & 200 x 200                                                       & 81                                                          & 810                                                     \\ \hline
Brodatz                                                    & \checkmark                                               &                                                                  & \checkmark                                                 & 32                                                        & 64 x 64                                                         & 64                                                          & 2048                                                    \\ \hline
CUReT                                                      & \checkmark                                               & \checkmark                                                       &                                                       & 61                                                        & 200 x 200                                                       & 92                                                          & 5612                                                    \\ \hline
\end{tabular}
}
\end{table}

\textbf{CUReT} database \cite{dana1999reflectance} includes 92 texture images per class and a total of 61 classes. This database is designed to contain large intra-class variation where images are captured under different illuminations and viewing directions with constant scale. All 92 images of 61 texture classes are cropped into 200 $\times$ 200 region and converted to gray scale \cite{varma2005statistical}.

\begin{figure}[htp]
%\captionsetup{justification=left}
\begin{minipage}[b]{1.0\linewidth}
\centering
\centerline{\includegraphics[clip=true, trim=0 320 05 170, width=0.99\linewidth]{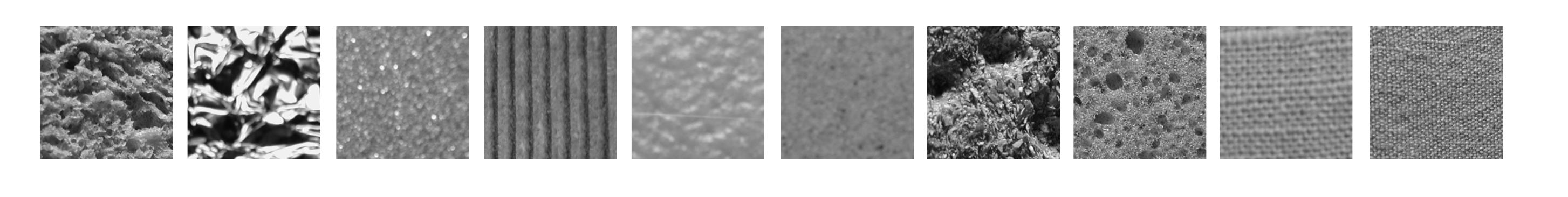}} %trim - left, bottom, right, top
\end{minipage}
\caption{Ten texture images randomly taken from each class of KTH-TIPS database.}
\label{fig:KTH_images}
\end{figure}

\textbf{KTH-TIPS} database \cite{hayman2004significance} is extended by imaging new samples of ten CUReT textures as shown in Fig.~\ref{fig:KTH_images}. 
It contains images with 3 different poses, 4 illuminations, and 9 different variations of scales with size $200 \times 200$. The \textbf{Brodatz}, \textbf{CUReT}, and \textbf{KTH-TIPS} databases are summarized in Table~\ref{tab:exp2}.

\subsection{Results of Experiment \#1}
\label{sub:result_ex1}
The results of Experiment \#1, accomplished on Outex\_TC10 and Outex\_TC12 are tabulated in Table~\ref{tab:com_ex1}. This table includes average classification accuracy of $K$-fold cross-validation test ($K$ = 10) for the proposed descriptor and the comparative summary of the results for $\textsc{Lbp}_{(R,N)}$~\cite{ojala2002multiresolution} , VZ-MR8~\cite{varma2005statistical}, VZ-Patch~\cite{varma2009statistical} and recent state-of-the-art methods. We have made the following observations from the results of Experiment \#1. Though the feature dimension \textsc{Lbpv} and \textsc{Lbp} are the same, due to additional contrast measure to the pattern histogram,  \textsc{Lbpv} produces a significantly better performance compared to the original \textsc{Lbp}. $\textsc{Lbp}_{R,N}^{riu2}/\textsc{Var}_{R,N}$ provides better performance compared to $\textsc{Lbpv}_{R,N}^{riu2}$. This is because \textsc{Lbp} and local variance of a texture image are complementary and so, the strength is produced by the joint distribution of \textsc{Lbp} and local variance compared to one alone. $\textsc{Clbp}\_S_{R,N}^{riu2}/M_{R,N}^{riu2}/C$(1,8 + 2,16 + 3, 24) which is made by fusing the $\textsc{Clbp}\_S$ and $\textsc{Clbp}\_M/C$, provides better performance compared to other variant of \textsc{Clbp}. This is because it contains complementary features of sign and magnitude, in addition to center pixel.
\begin{figure*}[htp]
%\captionsetup{justification=left}
\begin{minipage}[b]{1\linewidth}
\centering
\centerline{\includegraphics[clip=true, trim=10 315 10 30, width = 0.97\linewidth]{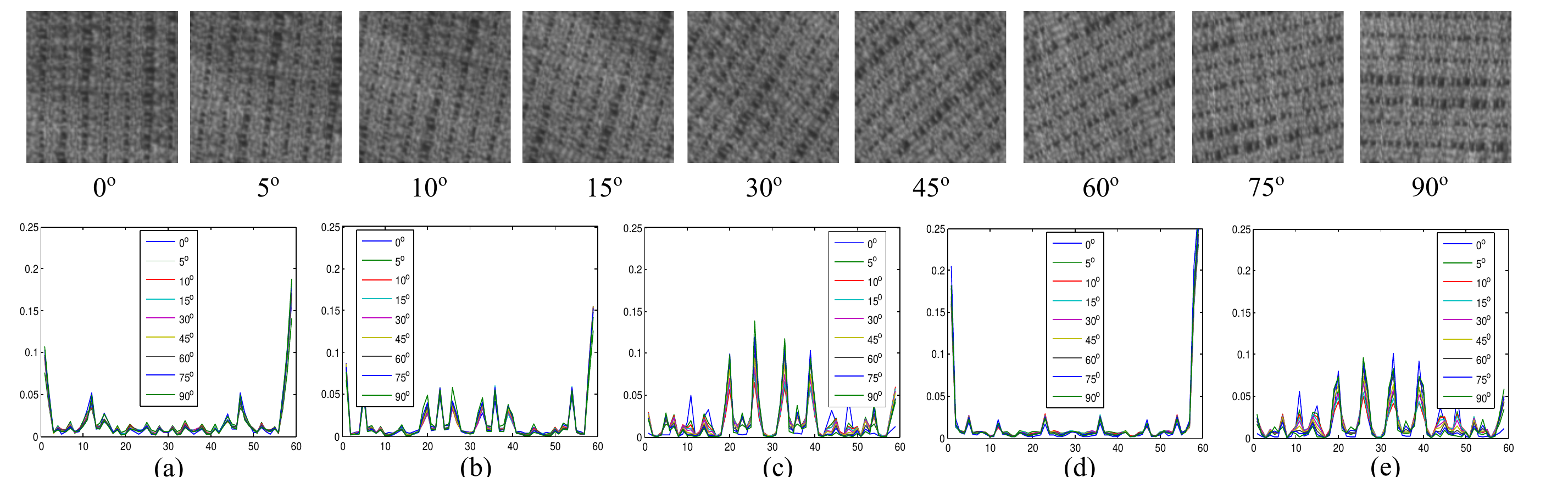}} %trim - left, bottom, right, top
\end{minipage}
\caption{(a)-(e) represent individual \textsc{Ljp} histograms $(H_{R, N}^{l}|l \in [2, \mathcal{L}])$ of a texture samples are taken from Outex database with 9 different orientations where abscissa and ordinate represent number of bins and feature probability distribution, respectively.}
\label{fig:TC12_o_r}
\end{figure*}
\begin{table}[htp]
\centering
\caption{Average classification accuracy (\%) on Outex\_TC10 and Outex\_TC12 using \textsc{State}-\textsc{Of}-\textsc{The}-\textsc{Art} schemes}
\label{tab:com_ex1}
\resizebox{\columnwidth}{!}{
\begin{tabular}{|c|c|c|c|c|c|}
\hline
\multirow{2}{*}{Methods} & \multirow{2}{*}{Classifier} & \multirow{2}{*}{Outex\_TC10} & \multicolumn{2}{c|}{Outex\_TC12} & \multirow{2}{*}{Average} \\ \cline{4-5}
                                  &                             &                              & horizon          & t184          &                          \\ \hline
\textsc{Ltp}~\cite{tan2007enhanced}                               & \textsc{Nnc}                         & 76.06                        & 63.42            & 62.56         & 67.34                    \\ \hline
\textsc{Var}~\cite{ojala1996comparative}                               & \textsc{Nnc}                         & 90.00                        & 64.35            & 62.93         & 72.42                    \\ \hline
\textsc{Lbp}~\cite{ojala2002multiresolution}                               & \textsc{Svm}                         &   97.60                           &              85.30    & 91.30              &                         91.40 \\ \hline
$\textsc{Lbp}_{R,N}^{riu2}$                               & \textsc{Nnc}                         & 84.89                             &  63.75                &              65.30 & 71.31                         \\ \hline
LBP/VAR                           & \textsc{Nnc}                         & 96.56                        & 78.08            & 79.31         & 84.65                    \\ \hline
$\textsc{Lbpv}_{R,N}^{riu2}$~\cite{guo2010rotation}                           & \textsc{Nnc}                         & 91.56                        & 77.01            & 76.62         & 81.73                    \\ \hline
CLBP\_S                           & \textsc{Nnc}                         & 84.81                        & 63.68            & 65.46         & 71.31                    \\ \hline
CLBP\_M                           & \textsc{Nnc}                         & 81.74                        & 62.77            & 59.30         & 67.93                    \\ \hline
CLBP\_M/C                         & \textsc{Nnc}                         & 90.36                        & 76.66            & 72.38         & 79.80                    \\ \hline
CLBP\_S\_M/C~\cite{guo2010completed}                      & \textsc{Nnc} & 94.53                        & 82.52            & 81.87         & 86.30                    \\ \hline
CLBP\_S/M                         & \textsc{Nnc}                         & 94.66                        & 83.14            & 82.75         & 86.85                    \\ \hline
CLBP\_S/M/C             & \textsc{Nnc}                         & 98.93                        & 92.29            & 90.30         & 93.84                    \\ \hline
$\textsc{Lbp}_{R, N}^{NT}$~\cite{fathi2012noise}                               & \textsc{Nnc}                         & 99.24                             &  96.18                &              94.28 & 96.56                         \\ \hline
$\textsc{Dlbp}_{R=3,N=24}$~\cite{liao2009dominant}                               & \textsc{Svm}                         & 98.10                             &  87.40                &              91.60 & 92.36                         \\ \hline
\textsc{Brint\_Cs\_Cm}~\cite{liu2014brint}                               & \textsc{Nnc}                         & 99.35                             &  97.69                &              98.56 & 98.53                         \\ \hline
VZ-MR8~\cite{varma2005statistical}                            & \textsc{Nsc}                             & 93.59                        & 92.82            & 92.55         & 92.99                    \\ \hline
VZ-Patch~\cite{varma2009statistical}                          &                         \textsc{Nsc}                             & 92.00                        & 92.06            & 91.41         & 91.82                    \\ \hline
$\textsc{Ptp}$~\cite{wang2013pixel}                               & \textsc{Nnc}                         & 99.56                             &  98.08                &              97.94 & 98.52                         \\ \hline
$\textsc{Cdcp}$~\cite{roy2017cdcp}                               & \textsc{Nnc}                         & 99.76                             &  99.82                &              99.62 & 99.72                         \\ \hline
Proposed \textsc{Ljp}                               & \textsc{Nnc}                         &  99.95                           &  99.97                & 99.81 &  99.91                        \\ \hline
Proposed \textsc{Ljp}                               & \textsc{Nsc}                         & 100                          &            99.97
 &    99.88            &  99.95                          \\ \hline
\end{tabular}}
\end{table}
\textsc{Dlbp} + \textsc{Ngf}, which makes use of the most frequently occurred 80\% patterns of \textsc{Lbp} to enhance the classification performance compared to the original $\textsc{Lbp}_{R,N}^{u2}$ but like $\textsc{Var}_{R,N}$ ignores the local spatial structure. In addition, the dimensionality of the \textsc{Dblp} varies with the training samples and it demands a pre-training stage. For comparison, we provide the best results of \textsc{Dlbp} with $R = 3$ and $N = 24$ in Table.~\ref{tab:com_ex1}. The state-of-the-art statistical algorithm, VZ-MR8 and VZ-Patch takes dense response from multiple filters. However, the performance is quite low compared to the proposed \textsc{Ljp}. In addition, the complexity of feature extraction and matching is quite high~\cite{varma2009statistical} compared to the proposed \textsc{Ljp} because the MR8 needs to calculate 8 maximum responses after 38 filters convolving with the image and use a clustering technique to build the textons dictionary. The $\textsc{Lbp}_{R,N}^{NT}$~\cite{fathi2012noise} based methods and \textsc{Brint}~\cite{liu2014brint} give better performance compared to other state-of-the-art \textsc{Lbp} methods. However, the accuracies are lower than those obtained by our proposed \textsc{Ljp}. This is mainly because $\textsc{Lbp}_{R,N}^{NT}$ extracts features by using locally rotation invariant $\textsc{Lbp}_{R,N}^{riu2}$ approach which produces only 10 bins and such small size of features can not represent each class well, while \textsc{Brint} extracted large number of features from multiple radius (R = 1, 2, 3, 4) by utilizing rotation invariant $\textsc{Lbp}^{ri}_{R, N}$ approach, whereas it loses the global image information.
% \begin{figure*}[htp]
% %\captionsetup{justification=left}
% \begin{minipage}[b]{1\linewidth}
% \centering
% \centerline{\includegraphics[clip=true, trim=10 315 10 30, width = 0.97\linewidth]{JET_S_R/Jet_Rot.pdf}} %trim - left, bottom, right, top
% \end{minipage}
% \caption{(a)-(e) represent individual \textsc{Ljp} histograms $(H_{R, N}^{l}|l \in [2, \mathcal{L}])$ of a texture samples are taken from Outex database with 9 different orientations where abscissa and ordinate represent number of bins and feature probability distribution, respectively.}
% \label{fig:TC12_o_r}
% \end{figure*}

Finally, even the Outex test suits (Outex\_TC10, Outex\_TC12 (``horizon'' and ``t184'')) contain both illuminant and rotation variant of textures the proposed descriptor achieves performance in term of mean accuracy and standard deviations 100 $\pm$ 0.0000\%, 99.97 $\pm$ 0.0732\% and 99.88 $\pm$ 0.1314\% on three different suits, respectively. Note that the textures under different illuminant and different viewpoint usually have different micro structures. The comparative results in Table~\ref{tab:com_ex1} shows that the proposed descriptor provides better classification performance compared to other state-of-the-art methods. The better performance is due to the following attributes. To extract \textsc{Ljp}, initially, a jet space representation of a texture image is derived from a set of derivative of Gaussian (DtGs) filter responses up to second order, so called local jet vector (\textsc{Ljv}) (Fig.~\ref{fig:PLJP}). Each element of \textsc{Ljv} is encoded via \textit{uniform} pattern scheme which produces 59 bins, are sufficient to discriminate each texture class. All the DtGs responses, \textsc{Ljv} together achieve invariance to scale, rotation, or reflection and are able to capture the \emph{micro structure} under different illumination controlled environment (``inca'', ``horizon'' and ``t184''). To visualize the rotation invariant characteristic of the proposed descriptor, an example of \textsc{Ljp} feature distribution for a texture samples taken from Outex\_TC10 database having 9 different orientations ($0^{\degree}, 5^{\degree}, 10^{\degree}, 15^{\degree}, 30^{\degree}, 45^{\degree}, 60^{\degree}, 75^{\degree}$, and $90^{\degree}$) are shown in Fig.~\ref{fig:TC12_o_r}. Fig~\ref{fig:TC12_o_r}(a)-(e) represent the histograms $(H_{R, N}^{l}|l \in [2, \mathcal{L}])$ of individual $(\textsc{Ljp}_{R,N}^{i,j,l}|l \in [2, \mathcal{L}])$. It is clearly observed that the \textsc{Ljp} feature distribution of different orientations are approximately overlapped which signify the rotation invariance of \textsc{Ljp} descriptor. 
\begin{figure*}[!htp]
%\captionsetup{justification=left}
\begin{minipage}[b]{1\linewidth}
\centering
\centerline{\includegraphics[clip=true, trim=10 250 10 80, width = 0.98\linewidth]{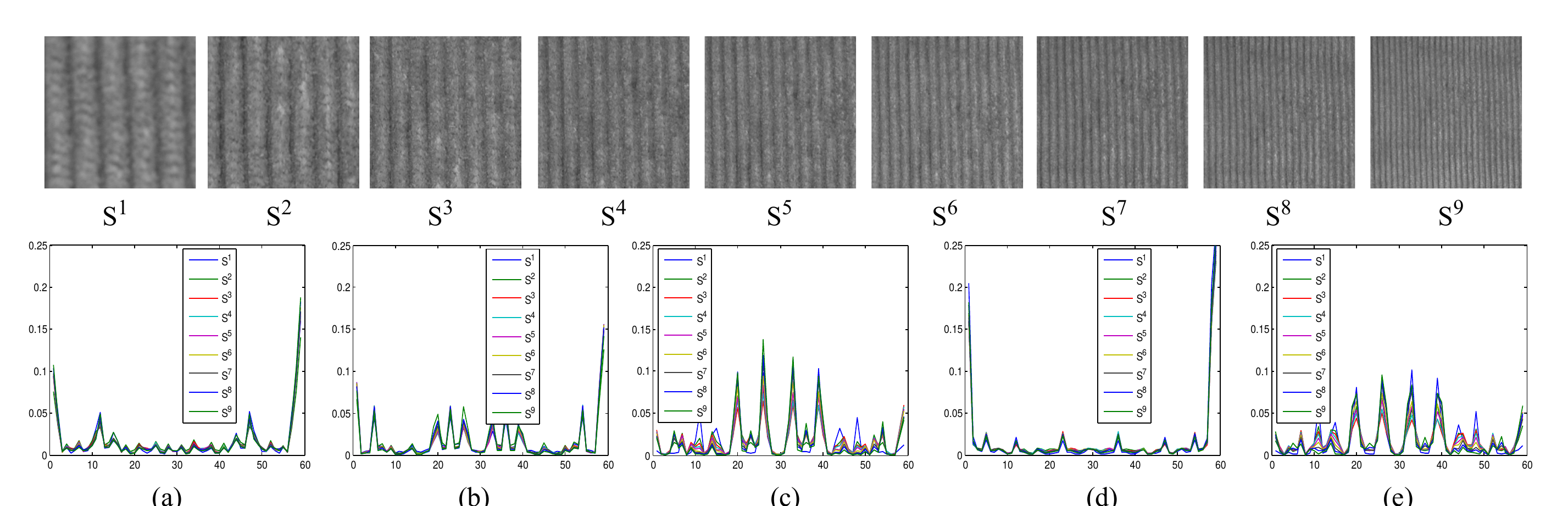}} %trim - left, bottom, right, top
\end{minipage}
\caption{(a)-(e) represent individual histograms $H_{R, N}^{l}$ of \textsc{Ljp}, where $R = 1, N = 8$ and $l \in [2, \mathcal{L}]$ of a texture sample are taken from KTH-TIPS~\cite{hayman2004significance} database having 9 different scales where abscissa and ordinate represent number of bins and feature probability distribution, respectively.}
\label{fig:KTH_lbp_ljp}
\end{figure*}

\subsection{Results of Experiment \#2}
\label{sub:result_ex2}
To analyse the scale invariance property of the proposed approach, we used the KTH-TIPS database which contains images with different scales. Each image of KTH-TIPS is captured in a controlled distance environment \cite{hayman2004significance}. Fig.~\ref{fig:KTH_lbp_ljp} shows an example texture image having $9$ different scales to illustrate the effectiveness of scale invariance properties of \textsc{Ljp} descriptor. Fig.~\ref{fig:KTH_lbp_ljp}($\mathrm{S^1}$)-($\mathrm{S^9}$) show nine images of ``corduroy" class in KTH-TIPS with different scales. Fig.~\ref{fig:KTH_lbp_ljp}(a)-(e) depict the histograms $(H_{R, N}^{l}|l \in [2, \mathcal{L}])$ of individual $\textsc{Ljp}_{R,N}^{i,j,l}$ descriptor with $R$ = 1, and $N$ = 8. It is clear from Fig.~\ref{fig:KTH_lbp_ljp}(a)-(e) that the histograms of individual \textsc{Ljp} for $9$ different scales are approximately closed which imply that the \textsc{Ljp} descriptor is scale invariant. The proposed descriptor achieves scale invariance by utilizing the properties of Hermite polynomials. As per Eqn.~(\ref{equ:HP}) the details within the Gaussian window is expanded over the basis of Hermite polynomials and this property provides a multi-scale hierarchical image structure for $\mathcal{L}$-jet although the scale ($\sigma$) is fixed~\cite{koenderink1987representation}.
% \begin{figure*}[!htp]
% %\captionsetup{justification=left}
% \begin{minipage}[b]{1\linewidth}
% \centering
% \centerline{\includegraphics[clip=true, trim=10 250 10 80, width = 0.98\linewidth]{JET_S_R/Jet_Scale.pdf}} %trim - left, bottom, right, top
% \end{minipage}
% \caption{(a)-(e) represent individual histograms $H_{R, N}^{l}$ of \textsc{Ljp}, where $R = 1, N = 8$ and $l \in [2, \mathcal{L}]$ of a texture sample are taken from KTH-TIPS~\cite{hayman2004significance} database having 9 different scales where abscissa and ordinate represent number of bins and feature probability distribution, respectively.}
% \label{fig:KTH_lbp_ljp}
% \end{figure*}

The average classification accuracy~(\%) of $K$-folds cross-validation test~($K$ = 10) for the proposed descriptor is shown in Table~\ref{tab:com_ex2}. This table also gives a comparative summary of the results for variants of \textsc{Lbp} and other state-of-the-art methods on three well known benchmark texture databases (KTH-TIPS, Brodatz, and CUReT). The following observations have been noted from Table~\ref{tab:com_ex2}. The proposed descriptor gives good performance with simple non-parametric \textsc{Nnc} classifier on three commonly used challenging databases. The performance is further improved, when an advanced classifier, \textsc{Nsc} is used. The \textsc{Nsc} classifier gives much better performance compared to the \textsc{Nnc} classifier for the same descriptor because: $(\romannum{1})$ for a given test sample \textsc{Nsc} uses all training samples of the same class while \textsc{Nnc} uses one training sample only;$(\romannum{2})$ \textsc{Nsc} is more robust to outlier and noise as it uses the correlation between different samples. \textsc{Dlbp} combining with Gabors feature, attains a higher classification rate than original \textsc{Lbp} with \textsc{Nnc}. However, its performance is quite less than the proposed \textsc{Ljp}. This is mainly because \textsc{Dlbp} does not consider scale variation, so it can't provide better performance on complex databases. The scale invariant feature $\textsc{Lbp}_{R,N}^{sri\_su2}$ provides better performance than $\textsc{Lbp}_{R, N}^{riu2}$ where $R$ = 1, 2, 3 and $N$ = 8, 16, 24, respectively. However the performance is lower than multi-resolution $\textsc{Lbp}_{R, N}^{riu2}$ and $\textsc{Clbp}\_S_{R,N}^{riu2}/M_{R,N}^{riu2}/C $, and much lesser than the proposed descriptor. This happens because the extraction of consistent and accurate scale for each pixel is difficult. The $\textsc{Lbp}_{R(i, j), 8}^{sri\_su2}$ provides good performance in controlled environment~\cite{li2012scale}, but it fails over more complex databases.
\begin{table}[htp]
\caption{Comparative results of classification accuracy achieved by the proposed and other state-of-the-art methods}
\begin{center}
\resizebox{\columnwidth}{!}{
\begin{tabular}{|c|c|c|c|c|}
\hline
\multirow{2}{*}{Methods} & \multirow{2}{*}{Classifier} & \multicolumn{3}{c|}{Classification Accuracy (\%)} \\ \cline{3-5} 
                         &                             & KTH-TIPS~\cite{hayman2004significance}  & Brodatz~\cite{brodatz1966textures}  & {CUReT~\cite{dana1999reflectance}} \\ \hline
                         
VZ-MR8~\cite{varma2005statistical} & \textsc{Nnc} & 94.50 & 94.62 & 97.43 \\ \hline

VZ-Patch~\cite{varma2009statistical} & \textsc{Nnc} & 92.40 & 87.10 & 98.03 \\ \hline
 
Lazebnik \textit{et al.}~\cite{lazebnik2005sparse} & \textsc{Nnc} & 91.30 & - & 72.50 \\ \hline
 
Zhang \textit{et al.}~\cite{zhang2007local} & \textsc{Svm} & 96.10 & - & 95.30 \\ \hline

\textsc{Mfs}~\cite{xu2006projective} & \textsc{Nnc}  & 81.62 & - & -\\ \hline

\textsc{Pfs}~\cite{quan2014distinct} & \textsc{Svm}  & 97.35 & - & -\\ \hline

\textsc{Bif}~\cite{crosier2010using} & Shift \textsc{Nnc}  & 98.50 & 98.47 & 98.60 \\ \hline

\textsc{Brint}~\cite{liu2014brint} & \textsc{Nnc}  & 97.75 & 99.22 & 97.06\\ \hline

$\textsc{Lbp}_{1, 8}^{riu2}$~\cite{ojala2002multiresolution} & \textsc{Nnc}  & 82.67 & 82.16 & 80.63 \\ \hline

$\textsc{Dlbp}_{3, 24}$~\cite{liao2009dominant} & \textsc{Svm}  & 86.99 & 99.16 & 84.93 \\ \hline

$\textsc{Lbp}_{1, 8}^{sri\_su2}$~\cite{li2012scale} & \textsc{Nnc}  & 89.73 & 69.50 & 85.00 \\ \hline

$\textsc{Lbp}_{(1,8 + 2,16 +3, 24)}$~\cite{ojala2002multiresolution} & \textsc{Nnc}  & 95.17 & - & 95.84 \\ \hline

$\textsc{Clbp\_smc}$~\cite{guo2010completed} & \textsc{Nnc}  & 97.19 & - & 97.40 \\ \hline

$\textsc{Sslbp}$~\cite{guo2016robust} & \textsc{Nnc}  & 97.80 & - & 98.55 \\ \hline

Proposed $\textsc{Ljp}$ & \textsc{Nnc}  & 98.12 & 97.31 & 98.12 \\ \hline

Proposed $\textsc{Ljp}$ & \textsc{Nsc}  & 99.75  &   99.16 & 99.65 \\ \hline

\end{tabular}
}
\end{center}
\label{tab:com_ex2}
\end{table}
Multi-scale \textsc{Bif}~\cite{crosier2010using} at scales $\sigma, 2\sigma, 4\sigma,$ and $8\sigma$ gives better performance than the proposed \textsc{Ljp} with \textsc{Nnc} classifier. This is mainly because \textsc{Bif} uses pyramid histogram with time inefficient shift matching scheme. However, the feature dimension of \textsc{Bif}~\cite{crosier2010using} is too large ($6^{4} = 1296$) compared to proposed \textsc{Ljp} ($59 \times 6 = 354$) and the shift matching scheme is time consuming. The performance of \textsc{Bif} reduces when scale shifting scheme is not considered~\cite{crosier2010using}.
\begin{table}[htp]
\caption{One way statistical ANOVA test results for Outex, KTH-TIPS, Brodatz, and CUReT databases, where level of significance is selected as $\alpha$ = 0.05.}
\begin{center}
\begin{tabular}{ c c c c c c } 
\hline
Source & SS & df & MS & F & Prob (p) $>$ F  \\ \hline
Groups & 2178.22 & 07 & 311.175 & 15.93 & $\mathbf{\mathrm{ 8.28653e^{-09}}}$\\ 
Error  & 0625.20 & 32 & 019.538 &   &   \\ 
Total  & 2803.42 & 39 &   &   &   \\ 
\hline
\end{tabular}
\end{center}
\label{tab:anova}
\end{table}

Although the images are captured under scale, rotations, and illumination variations, the proposed \textsc{Ljp} gives sound performance in term of mean accuracy and standard deviations i.e. 99.75 $\pm$ 0.790\%, 99.166 $\pm$ 0.4858\%, and 99.65 $\pm$ 0.45\%, on KTH-TIPS~\cite{hayman2004significance}, Brodatz~\cite{brodatz1966textures}, and CUReT~\cite{dana1999reflectance} texture databases~\cite{mikolajczyk2005performance}, respectively. The classification performance results in Table~\ref{tab:com_ex2} shows that the proposed descriptor achieves better and comparable performance compared to the state-of-the-art methods.
\begin{figure}[htp]
%\captionsetup{justification=left}
\begin{minipage}[b]{.99\linewidth}
\centering
\centerline{\includegraphics[clip=true, trim=120 300 140 300, width = 0.70\linewidth]{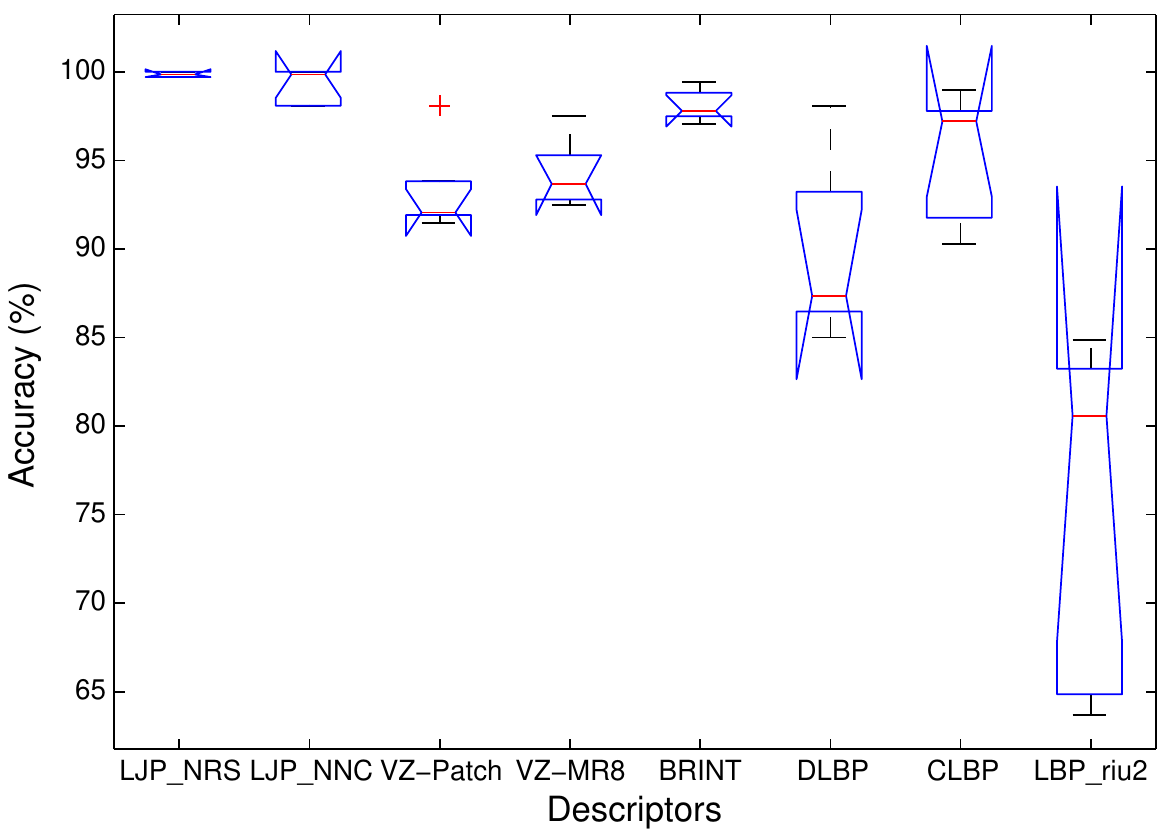}} %trim - left, bottom, right, top
\end{minipage}
\caption{The box plot (Descriptor vs. Accuracy) corresponding to one way statistical \textsc{Anova} test  for proposed \textsc{Ljp} and state-of-the-art descriptors on Outex, KTH-TIPS, Brodatz, CUReT databases.}
\label{fig:Anova}
\end{figure}
Though the trend is clear from the performance shown in Table~\ref{tab:com_ex1}, we have further analysed the performance using one way statistical Analysis of variance (\textsc{Anova}) test~\cite{iversen1987analysis}. \textsc{Anova} is a collection of statistical test used to analyze the differences among group means and their associated procedures. The null hypothesis $H_{0}$ for the test indicates that, \emph{there is no significant difference among group means}. We have taken the significant level $\alpha$ = 0.05 for this \textsc{Anova} test. We can reject $H_{0}$ if the $p$-value for an experiment is less than the selected significant level and which implies that the at least one group mean is significantly different from the others. To understand the performance of the proposed \textsc{Ljp} descriptor was significantly differs from well-known descriptors such as VZ-Patch, VZ-MR8, \textsc{Brint}, \textsc{Dlbp}, \textsc{Clbp}, and $\textsc{Lbp}^{riu2}$, we conduct an one way \textsc{Anova} test with significance level is kept as $\alpha$ = 0.05. The test results are shown in Table~\ref{tab:anova}. It is observed from Table~\ref{tab:anova} the $p$-value ($\mathbf{\mathrm{ 8.28653e^{-09}}}$) is less than the pre-select significant level $\alpha$ = 0.05 and this indicates that the performance of proposed descriptor significantly differs from other descriptors and hence reject the hypothesis $H_{0}$. In addition, the box plot corresponding to aforementioned \textsc{Anova} test is shown in Fig.~\ref{fig:Anova}, which also clearly indicates the mean performance of proposed descriptor is significantly better than the well-known descriptors such as VZ-MR8~\cite{varma2005statistical}, VZ-Patch~\cite{varma2009statistical}, \textsc{Brint}~\cite{liu2014brint}, \textsc{Dlbp}~\cite{liao2009dominant}, \textsc{Clbp}~\cite{guo2010completed}, and $\textsc{Lbp}^{riu2}$~\cite{ojala2002multiresolution}.

\subsection{Experiment \#3}
To study the classification performance of the proposed method in a noisy environment, the experiments are carried out on Outex\_TC10 texture database (discussed in subsec.~\ref{subsec:texture_db}) incorporating additive white Gaussian noise with different Signal to Noise Ratio~(\textsc{Snr}) in dB. Training and testing are done in the same situation as mentioned in noise-free condition.  
\begin{table}[htp]
\centering
\caption{Classification Accuracy (\%) of Proposed method and Different \textsc{State-Of-The-Art} methods on \textbf{Outex\_TC10} with Different Noise Levels in term of dB.}
\resizebox{\columnwidth}{!}{
\begin{tabular}{|c|c|c|c|c|c|c|}
\hline
\multirow{2}{*}{Methods} & \multirow{2}{*}{Classifier} & \multicolumn{5}{c|}{Classification Accuracy (\%)} \\ \cline{3-7} 
                         &                             & \textsc{Snr} = 100   & \textsc{Snr} = 30  & \textsc{Snr} = 15  & \textsc{Snr} = 10  & \textsc{Snr} = 5 \\ \hline
$\textsc{Lbp}_{R,N}^{riu2}$~\cite{ojala2002multiresolution}      &   \textsc{Nnc}           & 95.03          & 86.93        & 67.24         &  49.79       & 24.06       \\ \hline
$\textsc{Lbp}_{R,N,k}^{NT}$~\cite{fathi2012noise}      &   \textsc{Nnc}                      & -          &  99.79       & 99.76         &  99.76       &  99.74      \\ \hline
$\textsc{Clbp\_smc}$~\cite{guo2010completed}             &   \textsc{Nnc}                      & 99.30          & 98.12        & 94.58      & 86.07        & 51.22       \\ \hline
$\textsc{Ltp}_{R=3,N=24}^{riu2}$~\cite{tan2007enhanced}      &   \textsc{Nnc}                      & 99.45           & 98.31       & 93.44         & 84.32    & 57.37       \\ \hline
$\textsc{Nrlbp}_{R,N}^{riu2}$~\cite{ren2013noise}    &   \textsc{Nnc}                      & 84.49           & 81.16      & 77.52      & 70.16     & 50.88       \\ \hline
$\textsc{Brint}$~\cite{liu2014brint}                 &   \textsc{Nnc}                     & 97.76          & 96.48        &  95.47      & 92.97    & 88.31        \\ \hline
Proposed $\textsc{Ljp}_{R,N}^{l,u2}$                 &   \textsc{Nnc}                 & 99.95      & 99.83         & 99.79      &  99.74       & 99.74       \\ \hline
Proposed $\textsc{Ljp}_{R,N}^{l,u2}$            &   \textsc{Nsc}                 & 99.95      &  99.95         & 99.95     & 99.95     & 99.83       \\ \hline
\end{tabular}
}
\label{tab:com_ex3}
\end{table}
Table~\ref{tab:com_ex3} demonstrates the noise robustness of different methods on Outex\_TC10 database by comparing the classification rates for different noise levels~(measured using \textsc{Snr} i.e Signal to Noise Ratio). The proposed descriptor achieves state-of-the-art results in term of mean accuracy and standard deviations 99.95 $\pm$ 0.0976\%, 99.95 $\pm$ 0.0976\%, 99.95 $\pm$ 0.0976\%, 99.95 $\pm$ 0.0976\%, and 99.83 $\pm$ 0.0975\% on SNR = 100 dB, 30 dB, 15 dB, 10 dB, and 5 dB, respectively. It can be seen that only the proposed \textsc{Ljp} method is very modestly improved. This is due to the large smoothing kernels of DtGs which makes it robust to noise. The proposed method has inherited this property from the DtGs in contrast to the \textsc{Lbp} and its variants.
\begin{figure}[htp]
%\captionsetup{justification=left}
\begin{minipage}[b]{1.0\linewidth}
\centering
\centerline{\includegraphics[clip=true, trim=110 310 110 310, width=1.00\linewidth]{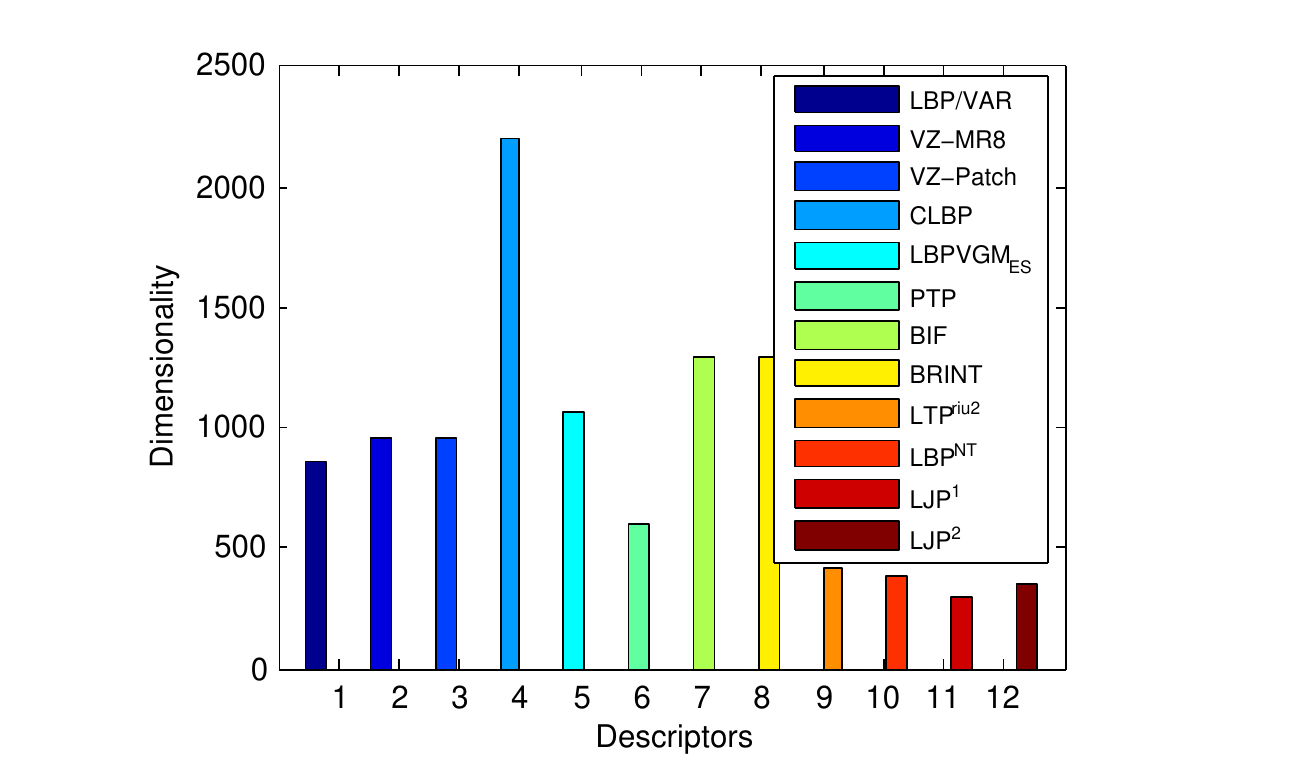}} %trim - left, bottom, right, top
\end{minipage}
\caption{Comparison of feature dimensionality between proposed and state-of-the-art features.}
\label{fig:class_acc_dim}
\end{figure}

The dimensionality of the proposed descriptors and other state-of-the-art descriptor are shown in Fig.~\ref{fig:class_acc_dim}. This figure clearly indicates that the dimensionality of the proposed descriptors ($\textsc{Ljp}^{1}$ has dimensionality of 295 and $\textsc{Ljp}^{2}$ has dimensionality 354) are significantly less than the other state-of-the-art methods. We have implemented the algorithm in \textsc{Matlab} 2011 environment and executed the program on $\mathrm{Intel^{\circledR}}$ $\mathrm{Core^{TM}}$2 Duo \textsc{Cpu} T6400 @ 2.00GHz $\times$ 2 processors and 3GB \textsc{Ram} with \textsc{Ubuntu} 14.04 \textsc{Lts} operating system. 
% The average time to extraction \textsc{Ljp} feature for a sample texture image of size (200 $\times$ 200) is approximately takes 0.17 seconds. Which shows that \textsc{Ljp} descriptor can be applicable in real time scenario.
The average feature extraction and matching time cost per-image on four benchmark texture databases by the proposed \textsc{Ljp} descriptor are listed in Table~\ref{tab:fea_ex_time}. It has been observed that the complexity of the proposed \textsc{Ljp} descriptor linearly varies with the image size and the matching complexity also varies linearly with the number of training samples. It shows that the proposed descriptor is fast enough for real time applications.
\begin{table}[htp]
\centering
\caption{Average feature extraction and matching time of the proposed method.}
\label{tab:fea_ex_time}
%\scalebox{0.85}{
\resizebox{\columnwidth}{!}{
\begin{tabular}{|c|c|c|c|c|c|}
\hline
\multirow{2}{*}{\begin{tabular}[c]{@{}c@{}} Time (sec)\end{tabular}} & \multicolumn{5}{c|}{Texture Databases}      \\ \cline{2-6} 
                                                                                         & KTH-TIPS & Brodatz & CUReT & Outex\_TC10 & Outex\_TC12 \\ \hline
\textsc{Ljp} Feature Extraction                                                                              &   0.22  &   0.115  &   0.190  &  0.119 &   0.122  \\ \hline
Matching using \textsc{Nnc}                                                                               &   0.0031  &  0.0080    &  0.0228 &  0.0164   &   0.0167   \\ \hline
Matching using \textsc{Nsc}                                                                              &   0.0109  &  0.0363   &  0.1432 &  0.0810   &   0.0839  \\ \hline
\end{tabular}
}
\end{table}

\section{Conclusions}
\label{sec:conclusion}
This paper proposes a simple, efficient yet robust descriptor, \textsc{Ljp} for texture classification. First, a jet space representation of a texture image is formed by the responses of a set of derivative of Gaussian~(DtGs) filter upto $2^{nd}$ order where DtGs responses preserves the intrinsic local image structure in a hierarchical way by utilizing the properties of Hermite polynomials. Then the local jet patterns~(\textsc{Ljp}) are computed by comparing the relation between center and neighboring pixels in the jet space. Finally, the histograms of \textsc{Ljp} for all elements of jet space representation are concatenated to form the feature vector. The results of the experimental study show that the proposed descriptor delivers quite auspicious performance under scale, rotation, reflection, and illumination variation and fairly robust to the noise. The comparative study indicates the proposed \textsc{Ljp} descriptor provides better performance compared to the state-of-the-art methods and in addition, the dimensionality of the final extracted feature vector is significantly less. While offering reasonably less time to extracts feature with a small feature dimension, the proposed \textsc{Ljp} can be applicable in real time scenario. 
\bibliographystyle{spmpsci}      % mathematics and physical sciences
\bibliography{reference_db}   % name your BibTeX data base

%\appendix

\begin{appendices}

\section{Effects of Similarity Transforms}
\label{subsec:EST}

In computer vision problems, a group of transforms are typically chosen to analyse the effect of geometrical structure those are invariant with respect to scaling, translation, rotation and reflection of an image, a constant intensity addition, and multiplication of image intensity by a positive factor and their combinations. In order to cope with the translation effect, we choose the analyzing point as the origin of the co-ordinate system.

In the jet space, Gaussian derivatives are represented by \textit{Hermite} polynomials multiplied with Gaussian window (in Eqn.~(\ref{equ:HP})). The details within the window is expanded over the basis of Hermite polynomials, therefore even if the scale ($\sigma$) is fixed, it provides a multi-scale hierarchical image structure for $\mathcal{L}$-jet~\cite{koenderink1987representation} 

Rotation of a surface is a more difficult problem in image analysis. Consider the rotation effect on jet about the origin with an angle $\theta$. Here the zero order derivative term of the jet ($J_{(0,0)}$) remains unaffected. The first order derivative terms are transformed as follows, 
$$\left(\begin{array}{cc} J_{(1,0)}\\J_{(0,1)} \end{array}\right) \to \left(\begin{array}{cc} \cos\theta & \sin\theta\\ -\sin\theta & \cos\theta \end{array}\right) \left(\begin{array}{cc} J_{(0,1)}\\J_{(1,0)} \end{array}\right)$$
and the second order terms are transformed according to
$$\left(\begin{array}{ccc} J_{(2,0)}\\J_{(1,1)}\\J_{(0,2)} \end{array}\right) \to \frac{1}{2} \left(\begin{array}{ccc} 1 + b & 2c & 1 - b\\ -c & 2b & c\\1 - b & -2c & 1 + b \end{array}\right) \left(\begin{array}{ccc} J_{(0,2)}\\J_{(1,1)}\\J_{(2,0)} \end{array}\right)$$

\noindent where $b = \cos2\theta$ and $c = \sin2\theta$. So, to return the starting values, the first order derivative structure requires a full $2\pi$ rotation, while the second order derivative structure returns after a rotation by $\pi$. 

In case of reflection (i.e. the effect of jet about the line $y = x$), the zero order term of the jet is not affected. However, the first order derivative term is transformed according to, $ J_{(1,0)} \longleftrightarrow J_{(0,1)}$ and the second order derivative term, according to $ J_{(2,0)} \longleftrightarrow J_{(0,2)}$. Considering all the DtGs responses together, the $\mathcal{L}$-jet~($\mathcal{L}$ = 6) achieves invariance to scale, rotation, or reflection.

Image structure analysis may be appropriate if it considers that intensities are non-negative unconstrained real numbers, and also do not exceed a maximum value. Therefore, the image structure should not be affected by adding a constant $(\alpha)$ to all intensities as in case of uniform illumination change. When such changes arise, only the zero order derivative term is affected and it simply transforms according to, $J_{(0,0)} \to J_{(0,0)} + \alpha$.

It is also required that the image local structure should be invariant to the multiplication by a non-zero positive factor ($\epsilon > 0$) with all image intensities. In this case, the factor is simply multiplied with all terms of jet vector i.e, $\vec{J} \to \epsilon\vec{J}$. Note that, it does not require multiplication of intensities by a negative factor to make it invariant to physical constrains.

\end{appendices}

%\begin{thebibliography}{}
% etc
%\end{thebibliography}

\end{document}